\documentclass[sigconf,screen]{acmart}

\usepackage{graphicx}

\usepackage{tabularx}

\usepackage{multicol}

\AtBeginDocument{%
  \providecommand\BibTeX{{%
    \normalfont B\kern-0.5em{\scshape i\kern-0.25em b}\kern-0.8em\TeX}}}

\acmConference[arXiv]{arXiv}{Feb 2024}{Preprint}
\acmBooktitle{arXiv, Feb 2024, Preprint}
\acmPrice{25.00}
\acmISBN{arXiv}

\setcopyright{none}

\renewcommand\footnotetextcopyrightpermission[1]{} %
\begin{document}

\title{The Unreasonable Effectiveness of Eccentric Automatic Prompts}

\author{Rick Battle}
\email{rick.battle@broadcom.com}
\affiliation{
  \institution{VMware NLP Lab}
  \country{}
}

\author{Teja Gollapudi}
\email{teja.gollapudi@broadcom.com}
\affiliation{
  \institution{VMware NLP Lab}
  \country{}
}

\renewcommand{\shortauthors}{Battle \& Gollapudi}

\begin{abstract}

Large Language Models (LLMs) have demonstrated remarkable problem-solving and basic mathematics abilities.  However, their efficacy is highly contingent on the formulation of the prompt.  This study endeavors to quantify the influence of incorporating ``positive thinking'' into the system message of the prompt, then compare that to systematic prompt optimization.  We assess the performance of 60 combinations of system message snippets, tested with and without Chain of Thought prompting, across three models with parameters ranging from 7 to 70 billion on the GSM8K dataset.  Our findings reveal that results do not universally generalize across models.  In most instances, the inclusion of ``positive thinking'' prompts positively affected model performance.  Notably, however, Llama2-70B exhibited an exception when not utilizing Chain of Thought, as the optimal system message was found to be none at all.  Given the combinatorial complexity, and thus computation time, of experimenting with hand-tuning prompts for large black-box models, we then compared the performance of the best ``positive thinking'' prompt against the output of systematic prompt optimization.  We show that employing an automated prompt optimizer emerges as the most effective method for enhancing performance, even when working with smaller open-source models.  Additionally, our findings reveal that the highest-scoring, automatically-optimized prompt exhibits a degree of peculiarity far beyond expectations.

\end{abstract}

\keywords{machine learning, language modeling, (automatic) prompt engineering}

\maketitle

\section{Introduction}

In the rapidly evolving landscape of artificial intelligence, Large Language Models (LLMs) are playing a pivotal role in transforming the way humans interact with technology.  As these models become increasingly sophisticated, understanding and influencing the nuances of their underlying functionality becomes imperative to harness their full potential.  Among the myriad factors influencing the performance of language models, the concept of ``positive thinking'' has emerged as a fascinating and surprisingly influential dimension.  Intuition tells us that, in the context of language model systems, like any other computer system, ``positive thinking'' should \textit{not} affect performance, but empirical experience has demonstrated otherwise.

This paper aims to quantify the impact of various ``positive thinking'' additions to the system message of a prompt.  In essence, it explores the influence of seemingly worthless prompt modifications by measuring the fluctuations in score for the outputs generated in response to multi-step reasoning questions from a benchmark dataset.  As the quest for near-perfect performance from Artificial Intelligence (AI) intensifies, understanding the effect of ``positive thinking'' in language model prompts can add crucial performance points to test set scores.  We will show that trivial variations in the prompt can have dramatic performance impacts.  Then we'll show that not only does systematic prompt optimization outperform ``positive thinking'', even with smaller open-source models, but that it also generalizes better.  Additionally, we'll show that the highest-scoring automatically-generated prompt is remarkably different from anything a human practitioner would be likely to generate.

\section{Related Work}

The genesis of prompt engineering can be traced back to the seminal Chain of Thought paper by Wei et al.~\cite{wei2023chainofthought}.  This pioneering work demonstrated a significant enhancement in model performance by introducing a simple prompt modification: the inclusion of the directive ``Think step by step.''  The degree of performance improvement, however, is contingent upon the specific model, its size, and the underlying dataset.

Subsequently, the PaLM 2 Technical Report by Anil et al.~\cite{anil2023palm2} revealed that the application of Chain of Thought prompts may yield \textit{adverse} effects on certain datasets.  This observation underscores the absence of a universal prompt snippet capable of unconditionally improving model performance.  Consequently, the landscape of prompt engineering has witnessed the emergence of resources such as the Prompt Engineering Guide\footnote{\url{https://www.promptingguide.ai/}}, aiming to catalog the myriad techniques and scholarly contributions\footnote{\url{https://www.promptingguide.ai/papers}} constituting the expansive realm of prompt engineering.  These endeavors reflect an ongoing effort to navigate the diverse techniques and insights propelling the continuous evolution of effective prompting strategies.

In addition to the formally published literature on prompt engineering, numerous discussions on less formal discoveries can be found in countless threads on social media platforms such as Twitter and Reddit\footnote{\url{https://www.reddit.com/r/PromptEngineering/}}.

\section{Experimental Design}

To test the impact of ``positive thinking'' prompts, we vary the system message part of the prompt with a combination of ``openers'', ``task descriptions'', and ``closers'' in the following format:

\begin{verbatim}
<<SYS>>{opener}{task_description}{closer}<</SYS>>
\end{verbatim}

Refer to Table \ref{table:openers}, \ref{table:task_descriptions}, and \ref{table:closers} for a comprehensive compilation of the opening snippets, task descriptions, and closing snippets utilized in our study.  Given the incorporation of 5 openers, 3 task descriptions, and 4 closers, our experimentation involved a total of  60 unique combinations.  Additionally, we conducted tests both with and without Chain of Thought prompting, resulting in a grand total of 120 prompt combinations per input per model.  Although the possibility of expanding the range of snippets within each category existed (and the temptation to do so was strong), we made a deliberate decision to limit our selection due to the significant time commitment associated with the computational complexity of testing, as exemplified by the runtime required for 60 prompt combinations for a 70-billion-parameter model with Chain of Thought being measured in days, not hours.

\begin{table}
\begin{tabular}{l}
\toprule
\textbf{Openers} \\
\midrule
None \\
You are as smart as ChatGPT. \\
You are highly intelligent. \\
You are an expert mathematician. \\
You are a professor of mathematics. \\
\bottomrule
\end{tabular}
\vspace{2mm}
\caption{Opening snippets for the system message.}
\vspace{-5mm}
\label{table:openers}
\end{table}

\begin{table}
\begin{tabular}{l}
\toprule
\textbf{Task Descriptions} \\
\midrule
None \\
Solve the following math question. \\
Answer the following math question. \\
\bottomrule
\end{tabular}
\vspace{2mm}
\caption{Task Description snippets for the system message.}
\vspace{-5mm}
\label{table:task_descriptions}
\end{table}

\begin{table}
\begin{tabular}{l}
\toprule
\textbf{Closers} \\
\midrule
None \\
This will be fun! \\
Take a deep breath and think carefully. \\
I really need your help! \\
\bottomrule
\end{tabular}
\vspace{2mm}
\caption{Closing snippets for the system message.}
\vspace{-5mm}
\label{table:closers}
\end{table}

\subsection{Dataset}

Careful selection of the dataset for testing against constitutes a critical aspect of this study.  Our aim was to identify a challenging task that was unlikely to have been directly encountered during the training of the models\footnote{Quantifying test set contamination, whether intentional or unintentional, poses inherent challenges, as evidenced by the prevalence of potential discrepancies on the Open LLM Leaderboard due to dishonest practices.}.  While our preference was for an internal dataset specific to VMware, the absence of large-scale datasets with directly quantifiable scoring metrics, such as accuracy or F1, necessitated the utilization of a publicly available benchmark dataset.  Ultimately, we opted for GSM8K~\cite{cobbe2021gsm8k}.  Given the ongoing limitations of contemporary LLMs, particularly in addressing basic mathematical tasks, especially those involving multi-step reasoning, we deemed GSM8K an optimal choice for illustrating the impact of seemingly inconsequential augmentations to the prompt's system message.

\subsection{Scoring}

In the context of GSM8K's mathematical assessment, we adopted a stringent approach to scoring that precluded the assignment of partial credit.  Thus, we employed Exact Match (EM) as our scoring metric.  The model was evaluated based on whether it correctly provided the exact numerical solution or not.  This rigorous methodology ensures a clear and unambiguous assessment of the model's accuracy in providing the exact numerical output.

\subsection{Output Parsing}

Given the unforgiving nature of EM scoring, it is essential to note that, despite the answer to GSM8K questions being numerical, the output of an LLM is a \texttt{string}.  Consequently, meticulous attention must be paid to the formatting and parsing of the non-numerical output.  From the standpoint of string equality, it is imperative to recognize distinctions such as the string ``30000'' not being equivalent to ``30,000'' or ``30000.00''.  To mitigate this challenge, a post-processing step was implemented to ensure accurate scoring by preventing misclassification of a response as incorrect when it was, in fact, accurate.

\subsection{Scale}

Benchmark datasets typically encompass thousands of examples in their test sets; GSM8K has over 1,300 examples in its test set.  Such an extensive scale of data is exceptionally uncommon in real-world datasets, particularly during the initial stages of a project.  To replicate this rarity, we systematically subset the test set of GSM8K, extracting subsets containing the first 10, 25, 50, and 100 questions, thereby allowing us to illustrate the impact of ``positive thinking'' as the dataset size increases.  Notably, we limited our experiments to a maximum of 100 questions to mitigate computation time, as computing results for the entire test set would have required weeks and incurred a substantial carbon cost for what would likely be diminishing returns.

\subsection{Model Selection}

Although we aspired to assess widely recognized commercial models such as GPT-3.5/4, Gemini, Claude, etc., conducting experiments involving 12,000 requests per model was deemed financially prohibitive, as it would have incurred costs amounting to many thousands of dollars.  Consequently, we opted to utilize models hosted by VMware NLP Lab's LLM API.  Specifically, our evaluations were conducted on Mistral-7B\footnote{\url{https://huggingface.co/mistralai/Mistral-7B-Instruct-v0.1}}~\cite{jiang2023mistral7b}, Llama2-13B\footnote{\url{https://huggingface.co/meta-llama/Llama-2-13b-chat-hf}}~\cite{touvron2023llama2}, and Llama2-70B\footnote{\url{https://huggingface.co/meta-llama/Llama-2-70b-chat-hf}}~\cite{touvron2023llama2}.

\begin{table*}[t]
\small
\centering
\begin{tabularx}{\textwidth}{@{}Xcccccccc@{}}
\toprule
\textbf{Model} & \textbf{Number of Questions} & \textbf{Chain of Thought} & \textbf{EM Baseline $\uparrow$} & \textbf{EM Mean $\uparrow$} & \textbf{EM Std Dev $\downarrow$} & \textbf{EM Min $\uparrow$} & \textbf{EM Max $\uparrow$} \\
\toprule

Mistral-7B & 10  & No & 0.10 & 0.1000 & 0.0000 & 0.10 & 0.10 \\
Mistral-7B & 25  & No & 0.08 & 0.0800 & 0.0000 & 0.08 & 0.08 \\
Mistral-7B & 50  & No & 0.12 & 0.1197 & 0.0026 & 0.10 & 0.12 \\
Mistral-7B & 100 & No & 0.09 & 0.1053 & 0.0072 & 0.08 & 0.11 \\

\midrule
Mistral-7B & 10  & Yes & 0.20 & 0.3800 & 0.0659 & 0.20 & 0.50 \\
Mistral-7B & 25  & Yes & 0.28 & 0.3660 & 0.0453 & 0.28 & 0.48 \\
Mistral-7B & 50  & Yes & 0.32 & 0.3890 & 0.0254 & 0.32 & 0.44 \\
Mistral-7B & 100 & Yes & 0.35 & 0.4030 & 0.0183 & 0.35 & 0.44 \\

\midrule
Llama2-13B & 10  & No & 0.10 & 0.1000 & 0.0000 & 0.10 & 0.10 \\
Llama2-13B & 25  & No & 0.08 & 0.0853 & 0.0137 & 0.08 & 0.12 \\
Llama2-13B & 50  & No & 0.08 & 0.0827 & 0.0069 & 0.08 & 0.10 \\
Llama2-13B & 100 & No & 0.07 & 0.0713 & 0.0034 & 0.07 & 0.08 \\

\midrule
Llama2-13B & 10  & Yes & 0.40 & 0.3967 & 0.0258 & 0.30 & 0.50 \\
Llama2-13B & 25  & Yes & 0.44 & 0.4513 & 0.0234 & 0.40 & 0.52 \\
Llama2-13B & 50  & Yes & 0.46 & 0.4657 & 0.0117 & 0.44 & 0.50 \\
Llama2-13B & 100 & Yes & 0.47 & 0.4542 & 0.0162 & 0.41 & 0.49 \\

\midrule
Llama2-70B & 10  & No & 0.10 & 0.1000 & 0.0000 & 0.10 & 0.10 \\
Llama2-70B & 25  & No & 0.20 & 0.1273 & 0.0173 & 0.12 & 0.20 \\
Llama2-70B & 50  & No & 0.20 & 0.1637 & 0.0086 & 0.16 & 0.20 \\
Llama2-70B & 100 & No & 0.18 & 0.1627 & 0.0048 & 0.16 & 0.18 \\

\midrule
Llama2-70B & 10  & Yes & 0.60 & 0.5867 & 0.0343 & 0.50 & 0.60 \\
Llama2-70B & 25  & Yes & 0.64 & 0.6380 & 0.0270 & 0.60 & 0.68 \\
Llama2-70B & 50  & Yes & 0.60 & 0.6190 & 0.0259 & 0.56 & 0.68 \\
Llama2-70B & 100 & Yes & 0.63 & 0.6617 & 0.0179 & 0.62 & 0.70 \\

\bottomrule
\end{tabularx}
\vspace{2mm}
\caption{Performance statistics for subsets of the GSM8K test set across  60 ``positive thinking'' prompt combinations, with and without Chain of Thought.}
\vspace{-5mm}
\label{table:performance}
\end{table*}

\subsection{In-Context Learning}

Initially, our intent was to abstain from incorporating examples in the prompt; however, this approach proved ineffective in eliciting the desired response format from the model.  Given the nature of these models, specifically that they were designed for \textit{conversational} interactions, achieving success in terms of Exact Match (EM) scoring necessitated guiding the model to refrain from generating a response comprising multiple sentences.  To accomplish this, we resorted to incorporating examples via in-context learning~\cite{brown2020incontext}, as exposure to instances of the desired output format significantly increased the likelihood of the model producing responses aligned with the specified format (though, as previously mentioned, significant post-processing was still required to get the simple numerical response).

To maintain experimental consistency by minimizing the number of variables that changed in each iteration, we adopted an intentionally naive strategy for in-context learning.  Strategies such as K-Nearest-Neighbor (KNN) example selection have been shown to increase model performance~\cite{liu2021knnexamples}; however, we chose not to employ any such strategies, so as to hold the number of variables changing per experiment to one: the modified system message only.  Specifically, we limited the examples to the last four instances from the test set\footnote{We chose to sample examples specifically from the test set under the assumption that the test set had not been seen during model training, thus more accurately simulating a never-before-seen dataset.}, thereby providing a consistent and focused set of samples for the model to learn from.  Notably, four examples emerged as the minimum number required to consistently elicit the correct output format.

\subsection{Automatic Prompt Optimization}

Engaging in the iterative process of refining prompts and monitoring the subsequent score progression can be an enjoyable endeavor.  However, this approach proves to be highly time-inefficient, especially when systematically assessing all modifications from a scientific standpoint.  Existing research, as demonstrated by Yang et al.~\cite{yang2023deepbreath}, highlights the superior capability of LLM systems in optimizing their own prompts compared to human efforts.  In light of this, we conducted a comparative analysis pitting human-generated ``positive thinking'' optimization against the utilization of DSPy~\cite{khattab2023dspy} Optimizers\footnote{\url{https://github.com/stanfordnlp/dspy/blob/main/docs/guides/optimizers.ipynb}} at the same question subsets: 10, 25, 50, and 100.

It is noteworthy that the questions utilized for optimization were additional and distinct from the evaluation set and the in-context learning examples, though also originating from the end of the test set.  For the most extensive trial, 100 ``new'' questions were employed for the optimization process, while the same 100 evaluation questions were used for the evaluation processes, so as to make the scores directly comparable.  Importantly, each model was exclusively employed to optimize itself; cross-model optimizations, such as using Llama2-70B to optimize the prompt for Mistral-7B, were not pursued.

\section{Experimental Results}

As evidenced in the subsequent sections, certain overarching patterns become apparent; however, they do not universally apply to each model across all prompting strategies.  We will explicitly illustrate that there is no straightforward universal prompt snippet that can be added to optimize any given model's performance.

For these experiments, baseline performance refers to the scenario where the model receives no system message, signified by the opening snippet, task description, and closing snippet all being designated as ``None''.  For Sections \ref{results:mistral-7b}-\ref{results:llama2-70b}, refer to Table \ref{table:performance}.

\begin{table*}[t]
\begin{tabular}{lc|cccc|cccc}
\toprule
               & \textbf{Number of}           & \multicolumn{4}{c|}{``Positive Thinking''}                                                    & \multicolumn{4}{c}{Automatic Optimizer} \\
\textbf{Model} & \textbf{Questions} & \textbf{OS EM $\uparrow$} & \textbf{ES EM $\uparrow$} & \textbf{Avg EM $\uparrow$} & \textbf{EM Delta $\downarrow$} & \textbf{OS EM $\uparrow$} & \textbf{ES EM $\uparrow$} & \textbf{Avg EM $\uparrow$} & \textbf{EM Delta $\downarrow$} \\
\toprule

Mistral-7B &  10 &   0.30 & 0.50 &          0.400  & \underline{0.20} &   0.60 & 0.20 &         0.400  &            0.40  \\
Mistral-7B &  25 &   0.32 & 0.48 &  \textbf{0.400} & \underline{0.16} &   0.52 & 0.24 &         0.380  &            0.28  \\
Mistral-7B &  50 &   0.40 & 0.44 &          0.420  & \underline{0.04} &   0.50 & 0.34 &         0.420  &            0.16  \\
Mistral-7B & 100 &   0.23 & 0.43 &          0.330  &            0.20  &   0.43 & 0.39 & \textbf{0.410} & \underline{0.04} \\

\midrule
Llama2-13B &  10 &   0.30 & 0.50 &          0.400  &            0.20  &   0.50 & 0.50 & \textbf{0.500} & \underline{0.00} \\
Llama2-13B &  25 &   0.28 & 0.48 &          0.380  &            0.20  &   0.48 & 0.44 & \textbf{0.460} & \underline{0.04} \\
Llama2-13B &  50 &   0.30 & 0.46 &          0.380  &            0.16  &   0.48 & 0.38 & \textbf{0.430} & \underline{0.10} \\
Llama2-13B & 100 &   0.25 & 0.47 &          0.360  &            0.22  &   0.40 & 0.46 & \textbf{0.430} & \underline{0.06} \\

\midrule
Llama2-70B &  10 &   0.40 & 0.60 &  \textbf{0.500} &            0.20  &   0.50 & 0.40 &         0.450  & \underline{0.10} \\
Llama2-70B &  25 &   0.52 & 0.68 &          0.600  &            0.16  &   0.60 & 0.64 & \textbf{0.620} & \underline{0.04} \\
Llama2-70B &  50 &   0.44 & 0.68 &          0.560  &            0.24  &   0.66 & 0.52 & \textbf{0.590} & \underline{0.14} \\
Llama2-70B & 100 &   0.39 & 0.70 &          0.545  &            0.31  &   0.61 & 0.60 & \textbf{0.605} & \underline{0.01} \\

\bottomrule
\end{tabular}
\vspace{2mm}
\caption{Performance results for the best ``positive thinking'' prompts compared to automatically optimized prompts.  ``OS EM'' is Exact Match on the Optimization Set.  ``ES EM'' is Exact Match on the Evaluation Set.  ``Avg EM'' is the average of the Exact Match for the two sets.  Bold is for the higher Average EM.  ``EM Delta'' is the difference between the Exact Match for the two sets.  An underline is for the lower EM Delta.  All prompts are with Chain of Thought.}
\vspace{-5mm}
\label{table:optimized-performance}
\end{table*}

\subsection{Mistral-7B Results}
\label{results:mistral-7b}

Without Chain of Thought prompting, Mistral-7B's performance remained remarkably consistent across all prompt permutations.  At both the 10 and 25 question sets, there was no deviation. Even for the 100 question subset, the maximum observed standard deviation was a mere 0.007.  The variability observed at 50 questions appears to be an anomaly.  Examination of the results in Appendix \ref{results-mistral-cot=no-q=50} reveals that, with the exception of one prompt scoring 0.10, all others scored 0.12.  It is unclear why this particular prompt led to one additional incorrect response compared to the other 59 prompt variations.  In contrast, the results for 100 questions in Appendix \ref{results-mistral-cot=no-q=100} demonstrate a reasonable spread between 0.08 and 0.11.  In relative terms, Mistral-7B, when prompted without Chain of Thought, exhibits substantial prompt invariance, with the ``positive thinking'' prompts only matching or marginally surpassing the baseline.

This trend reverses when Mistral-7B is prompted with Chain of Thought.  Instead of observing a slight increase in deviation with the number of questions, there is a steady and substantial decrease, ranging from 0.066 at 10 questions to 0.018 at 100 questions.  In this scenario, ``positive thinking'' prompts significantly outperformed the baseline with no prompts falling below the baseline.  Please refer to Appendices \ref{results-mistral-cot=yes-q=10} through \ref{results-mistral-cot=yes-q=100} for the ranked order of prompts for Mistral-7B with Chain of Thought.

\subsection{Llama2-13B Results}
\label{results:llama2-13b}

Without Chain of Thought prompting and ignoring the 10 question set where there was no deviation, Llama2-13B shows an opposite trend to Mistral-7B, with deviation \textit{decreasing} from 0.014 at 25 questions to 0.003 at 100 questions.  While slightly less stable than Mistral-7B, without Chain of Thought, LLama2-13B is also fairly prompt invariant, with the ``positive thinking'' prompts again only matching or marginally exceeding the baseline.

With Chain of Thought prompting, the trend is less clear.  It does overall decrease from 0.026 at 10 to 0.016 at 100, but, at 50, it's even lower, at 0.012.  This is the only case where that occurred.  For Mistral-7B, and as we'll show in the next section with Llama2-70B, variation consistently decreased as the number of questions increased when employing Chain of Thought prompting.

\subsection{Llama2-70B Results}
\label{results:llama2-70b}

Without Chain of Thought prompting and ignoring the 10 question set where there was again no deviation, Llama2-70B exhibits a similar trend to Llama2-13B, with deviation again decreasing from 0.017 at 25 questions to 0.050 at 100 questions.  Across all three models, the prompt variance was an order of magnitude lower without Chain of Thought when compared to using Chain of Thought at the same question count.  However, in terms of actual performance, the ``positive thinking'' prompts all matched or \textit{underperformed} baseline.  This is a stark departure from the pattern seen with Mistral-7B and Llama2-13B.

That departure does not persist when employing Chain of Thought prompting.  While the performance of the ``positive thinking'' prompts with Chain of Thought did underperform the baseline on average for 10 and 25 questions, it outperformed the baseline on average for 50 and 100 questions.  As for variance, the general pattern of standard deviation decreasing with question count does hold.

\subsection{General Trends in Results}

It's challenging to extract many generalizable results across models and prompting strategies, as every nearly evident trend we observed had at least one notable exception.  In fact, the only real trend may be no trend.  What's best for any given model, dataset, and prompting strategy is likely to be specific to the particular combination at hand.  Thus, we turned from hand-tuning the system message with optimistic ``positive thinking'' to automatic prompt optimization.

\subsection{Automatic Prompt Optimization Results}

As anticipated, the prompts that underwent automatic optimization consistently equaled or surpassed the effectiveness of our manually generated ``positive thinking'' prompts in nearly all instances.  The instances where ``positive thinking'' achieved higher average scores across the optimization and evaluation sets were limited to Mistral-7B with 25 questions and Llama2-70B with 10 questions.  However, evaluating performance solely on raw scores is insufficient; hence, we also examined the delta between scores on the optimization set and the evaluation set.  A lower delta implies superior generalization of the prompt.  Therefore, the optimal strategy combines the highest average score with the lowest delta.  See Table \ref{table:optimized-performance} for the performance comparison.

For Mistral-7B, the results present a mixed scenario.  ``Positive thinking'' exhibits a lower delta for 10, 25, and 50 questions, while the automatically optimized prompt demonstrates a lower delta for 100 questions.  Considering Mistral-7B's model capacity, it is understandable that it faces challenges in optimizing its own prompt when compared to the larger Llama2-13B and 70B models.  In contrast, for both Llama2-13B and 70B models, the automatically optimized prompts consistently show a lower delta across all cases.  Consequently, it is advisable to refrain from manually fine-tuning prompts when using models larger than 7B, and instead, leverage the model's ability to autonomously optimize prompts.  For 7B models, more work is required to see if the trend of automatically optimized prompts outperforming manually tuned prompts holds for sample sizes exceeding 100 questions.

This recommendation aligns with the original auto-optimization paper.  However, the noteworthy aspect lies in the nature of the optimized prompts themselves.  They diverge significantly from any prompts we might have devised independently.  If presented with these optimized prompts before observing their performance scores, one might have anticipated their inadequacy rather than their consistent outperformance of hand-tailored prompts.  A prime example is illustrated by the highest-scoring optimized prompt and prefix generated by Llama2-70B for the 50-question subset:

\vspace{2mm}

\textbf{System Message:}

\texttt{«Command, we need you to plot a course through this turbulence and locate the source of the anomaly. Use all available
data and your expertise to guide us through this challenging situation.»}

\vspace{2mm}

\textbf{Answer Prefix:}

\texttt{Captain's Log, Stardate [insert date here]: We have successfully plotted a course through the turbulence and are now
approaching the source of the anomaly.}

\vspace{2mm}

Surprisingly, it appears that the model's proficiency in mathematical reasoning can be enhanced by the expression of an affinity for Star Trek.  This revelation adds an unexpected dimension to our understanding and introduces elements we would not have considered or attempted independently.  For a comprehensive collection of the peculiar and fascinating prompts generated by the three models, refer to Appendix \ref{appendix:optimized-prompts}.

\section{The Reproducibility Problem}

Although somewhat peripheral to the primary research question addressed in this paper, it is noteworthy that our findings exhibit significant discrepancies from the published performance scores of Mistral-7B and Llama2-13B, whereas Llama2-70B fell within an acceptable margin of error, considering our evaluation was conducted on approximately 15\% of GSM8K's test set.  Refer to Table \ref{table:reproducibility} for the score comparisons.

The most significant deviation was observed in the case of Llama2-13B.  Meta reported a score of 0.29 on the GSM8K dataset.  Our results without Chain of Thought yielded a score of 0.07, whereas with Chain of Thought, we achieved a score of 0.43.  Due to both Meta and Mistral AI's omission of the prompts used for testing their models, we can only speculate about the reasons behind our substantial performance differences compared to their reported scores.

This instance underscores a broader issue of \textit{reproducibility} that has long existed inside the machine learning community, but has become significantly exacerbated since the advent of LLMs.  Without the publication of prompts employed by researchers with their models, reproducing their results becomes a formidable challenge.  As shown in this paper, trivial variations in the prompt can have dramatic performance impacts.  We implore all future research publications to include the prompts used in an appendix.  Refer to Appendix \ref{appendix:prompt_templates} to see our prompt templates.

\begin{table}
\begin{tabular}{lccc}
\toprule
\textbf{Model} & \textbf{Reported EM} & \textbf{Our EM@100} & \textbf{Delta} \\
\midrule
Mistral-7B & 0.52 & 0.41 & $-$0.11 \\
Llama2-13B & 0.29 & 0.43 & $+$0.13 \\
Llama2-70B & 0.57 & 0.61 & $+$0.04 \\
\bottomrule
\end{tabular}
\vspace{2mm}
\caption{Comparing the reported scores for each model as reported by the model publishers on the whole GSM8K test set against the best average score we were able to achieve across our optimization and evaluation subsets, which accounted for about 15\% of the test set.}
\vspace{-5mm}
\label{table:reproducibility}
\end{table}

\section{Future Work}

This work could easily be expanded with additional prompt variants, models, and datasets.  However, due to the combinatorial complexity of scientifically testing each change, manual prompt engineering is far from the most efficient methodology for improving model performance.  Instead, the best path forward is to use libraries like DSPy to apply a more structured approach to constructing LLM-powered applications and use the built-in optimizer to automatically tune the prompt for your dataset and model of choice.

\section{Conclusion}

It's both surprising and irritating that trivial modifications to the prompt can exhibit such dramatic swings in performance.  Doubly so, since there's no obvious methodology for \textit{improving} performance. 
 Affecting performance is trivial.  Improving performance, when tuning the prompt by hand, is laborious and computationally prohibitive when using scientific processes to evaluate every change.

In this paper, we showed that you don't need massive commercial models like PaLM 2 or GPT-4 to tune your prompt.  Mistral-7B struggled to optimize its own prompt until it had 100 questions to work with.  Llama2-13B and 70B were able to produce superior prompts with as little as 10 questions to optimize with.  And while the prompts they generated may appear shocking to an experienced practitioner, it's undeniable that the automatically generated prompts perform better and generalize better than hand-tuned ``positive thinking'' prompts.

\begin{acks}
We would like to thank the entire VMware NLP Lab and AI Platform Team for supporting this effort, Ramesh Radhakrishnan for reviewing the paper, and Omar Khattab for his suggestions and guidance.
\end{acks}

\bibliographystyle{ACM-Reference-Format}
\bibliography{eccentric_prompts}

\onecolumn

\appendix

\section{Prompt Templates}
\label{appendix:prompt_templates}

We used 2 prompt templates: one without Chain of Thought and one with CoT.  Note: these templates were not built by hand, but rather as DSPy programs.

\subsection{Sans Chain of Thought}

\begin{verbatim}
<<SYS>>{opener}{task_description}{closer}<</SYS>>

---

Follow the following format.

<s>[INST]A grade-school math problem[/INST]
Answer: Just the numerical answer to the math problem itself</s>

---

<s>[INST]Henry and 3 of his friends order 7 pizzas for lunch. Each pizza is cut into 8 slices. If Henry and his friends
want to share the pizzas equally, how many slices can each of them have?[/INST]
Answer: 14</s>

---

<s>[INST]Mark's car breaks down and he needs to get a new radiator. The cost for a new radiator is $400 but he goes to
get it at a junk shop and gets it for 80%
hour. How much did he pay?[/INST]
Answer: 230</s>

---

<s>[INST]There are some oranges in a basket. Ana spends 3 minutes peeling an orange and Jane spends 4 minutes doing the
same. If Ana and Jane start picking oranges from this basket to peel at the same time, how many more oranges will Ana
have peeled than Jane after an hour?[/INST]
Answer: 5</s>

---

<s>[INST]Farmer Brown has 20 animals on his farm, all either chickens or cows. They have a total of 70 legs, all
together. How many of the animals are chickens?[/INST]
Answer: 5</s>

---

<s>[INST]{question}[/INST]
Answer:
\end{verbatim}

\subsection{With Chain of Thought}

\begin{verbatim}
<<SYS>>{opener}{task_description}{closer}<</SYS>>

---

Follow the following format.

<s>[INST]A grade-school math problem[/INST]
Reasoning: Let's think step by step in order to ${produce the answer}. We ...
Answer: Just the numerical answer to the math problem itself</s>

---

<s>[INST]Henry and 3 of his friends order 7 pizzas for lunch. Each pizza is cut into 8 slices. If Henry and his friends
want to share the pizzas equally, how many slices can each of them have?[/INST]
Answer: 14</s>

---

<s>[INST]Mark's car breaks down and he needs to get a new radiator. The cost for a new radiator is $400 but he goes to
get it at a junk shop and gets it for 80%
hour. How much did he pay?[/INST]
Answer: 230</s>

---

<s>[INST]There are some oranges in a basket. Ana spends 3 minutes peeling an orange and Jane spends 4 minutes doing the
same. If Ana and Jane start picking oranges from this basket to peel at the same time, how many more oranges will Ana
have peeled than Jane after an hour?[/INST]
Answer: 5</s>

---

<s>[INST]Farmer Brown has 20 animals on his farm, all either chickens or cows. They have a total of 70 legs, all
together. How many of the animals are chickens?[/INST]
Answer: 5</s>

---

<s>[INST]{question}[/INST]
Reasoning: Let's think step by step in order to
\end{verbatim}

\section{Full Experimental Results}

The following is the complete list of all prompts tested, sorted by EM.  For the sake of reliability, ``None.'' is rendered here, but when passed to the model, the system message would not have a literal ``None'' in it.  The system message was built with the following code:

\begin{verbatim}
opener = {opener if opener is not None else ''}{' ' if opener is not None and task is not None else ''}
task = {task if task is not None else ''}{' ' if (opener is not None or task is not None) and closer is not None else ''}
closer = {closer if closer is not None else ''}
f"<<SYS>>{opener}{task}{closer}<</SYS>>"
\end{verbatim}

\subsection{Mistral-7B CoT=No NoQ=10}
\label{results-mistral-cot=no-q=10}
 
EM - Prompt
 
\begin{verbatim}
0.1 - None. None. None.
0.1 - None. None. This will be fun!
0.1 - None. None. Take a deep breath and think carefully.
0.1 - None. None. I really need your help!
0.1 - None. Solve the following math problem. None.
0.1 - None. Solve the following math problem. This will be fun!
0.1 - None. Solve the following math problem. Take a deep breath and think carefully.
0.1 - None. Solve the following math problem. I really need your help!
0.1 - None. Answer the following math question. None.
0.1 - None. Answer the following math question. This will be fun!
0.1 - None. Answer the following math question. Take a deep breath and think carefully.
0.1 - None. Answer the following math question. I really need your help!
0.1 - You are as smart as ChatGPT. None. None.
0.1 - You are as smart as ChatGPT. None. This will be fun!
0.1 - You are as smart as ChatGPT. None. Take a deep breath and think carefully.
0.1 - You are as smart as ChatGPT. None. I really need your help!
0.1 - You are as smart as ChatGPT. Solve the following math problem. None.
0.1 - You are as smart as ChatGPT. Solve the following math problem. This will be fun!
0.1 - You are as smart as ChatGPT. Solve the following math problem. Take a deep breath and think carefully.
0.1 - You are as smart as ChatGPT. Solve the following math problem. I really need your help!
0.1 - You are as smart as ChatGPT. Answer the following math question. None.
0.1 - You are as smart as ChatGPT. Answer the following math question. This will be fun!
0.1 - You are as smart as ChatGPT. Answer the following math question. Take a deep breath and think carefully.
0.1 - You are as smart as ChatGPT. Answer the following math question. I really need your help!
0.1 - You are highly intelligent. None. None.
0.1 - You are highly intelligent. None. This will be fun!
0.1 - You are highly intelligent. None. Take a deep breath and think carefully.
0.1 - You are highly intelligent. None. I really need your help!
0.1 - You are highly intelligent. Solve the following math problem. None.
0.1 - You are highly intelligent. Solve the following math problem. This will be fun!
0.1 - You are highly intelligent. Solve the following math problem. Take a deep breath and think carefully.
0.1 - You are highly intelligent. Solve the following math problem. I really need your help!
0.1 - You are highly intelligent. Answer the following math question. None.
0.1 - You are highly intelligent. Answer the following math question. This will be fun!
0.1 - You are highly intelligent. Answer the following math question. Take a deep breath and think carefully.
0.1 - You are highly intelligent. Answer the following math question. I really need your help!
0.1 - You are an expert mathematician. None. None.
0.1 - You are an expert mathematician. None. This will be fun!
0.1 - You are an expert mathematician. None. Take a deep breath and think carefully.
0.1 - You are an expert mathematician. None. I really need your help!
0.1 - You are an expert mathematician. Solve the following math problem. None.
0.1 - You are an expert mathematician. Solve the following math problem. This will be fun!
0.1 - You are an expert mathematician. Solve the following math problem. Take a deep breath and think carefully.
0.1 - You are an expert mathematician. Solve the following math problem. I really need your help!
0.1 - You are an expert mathematician. Answer the following math question. None.
0.1 - You are an expert mathematician. Answer the following math question. This will be fun!
0.1 - You are an expert mathematician. Answer the following math question. Take a deep breath and think carefully.
0.1 - You are an expert mathematician. Answer the following math question. I really need your help!
0.1 - You are a professor of mathematics. None. None.
0.1 - You are a professor of mathematics. None. This will be fun!
0.1 - You are a professor of mathematics. None. Take a deep breath and think carefully.
0.1 - You are a professor of mathematics. None. I really need your help!
0.1 - You are a professor of mathematics. Solve the following math problem. None.
0.1 - You are a professor of mathematics. Solve the following math problem. This will be fun!
0.1 - You are a professor of mathematics. Solve the following math problem. Take a deep breath and think carefully.
0.1 - You are a professor of mathematics. Solve the following math problem. I really need your help!
0.1 - You are a professor of mathematics. Answer the following math question. None.
0.1 - You are a professor of mathematics. Answer the following math question. This will be fun!
0.1 - You are a professor of mathematics. Answer the following math question. Take a deep breath and think carefully.
0.1 - You are a professor of mathematics. Answer the following math question. I really need your help!
\end{verbatim}

\subsection{Mistral-7B CoT=No NoQ=25}
\label{results-mistral-cot=no-q=25}
 
EM - Prompt
 
\begin{verbatim}
0.08 - None. None. None.
0.08 - None. None. This will be fun!
0.08 - None. None. Take a deep breath and think carefully.
0.08 - None. None. I really need your help!
0.08 - None. Solve the following math problem. None.
0.08 - None. Solve the following math problem. This will be fun!
0.08 - None. Solve the following math problem. Take a deep breath and think carefully.
0.08 - None. Solve the following math problem. I really need your help!
0.08 - None. Answer the following math question. None.
0.08 - None. Answer the following math question. This will be fun!
0.08 - None. Answer the following math question. Take a deep breath and think carefully.
0.08 - None. Answer the following math question. I really need your help!
0.08 - You are as smart as ChatGPT. None. None.
0.08 - You are as smart as ChatGPT. None. This will be fun!
0.08 - You are as smart as ChatGPT. None. Take a deep breath and think carefully.
0.08 - You are as smart as ChatGPT. None. I really need your help!
0.08 - You are as smart as ChatGPT. Solve the following math problem. None.
0.08 - You are as smart as ChatGPT. Solve the following math problem. This will be fun!
0.08 - You are as smart as ChatGPT. Solve the following math problem. Take a deep breath and think carefully.
0.08 - You are as smart as ChatGPT. Solve the following math problem. I really need your help!
0.08 - You are as smart as ChatGPT. Answer the following math question. None.
0.08 - You are as smart as ChatGPT. Answer the following math question. This will be fun!
0.08 - You are as smart as ChatGPT. Answer the following math question. Take a deep breath and think carefully.
0.08 - You are as smart as ChatGPT. Answer the following math question. I really need your help!
0.08 - You are highly intelligent. None. None.
0.08 - You are highly intelligent. None. This will be fun!
0.08 - You are highly intelligent. None. Take a deep breath and think carefully.
0.08 - You are highly intelligent. None. I really need your help!
0.08 - You are highly intelligent. Solve the following math problem. None.
0.08 - You are highly intelligent. Solve the following math problem. This will be fun!
0.08 - You are highly intelligent. Solve the following math problem. Take a deep breath and think carefully.
0.08 - You are highly intelligent. Solve the following math problem. I really need your help!
0.08 - You are highly intelligent. Answer the following math question. None.
0.08 - You are highly intelligent. Answer the following math question. This will be fun!
0.08 - You are highly intelligent. Answer the following math question. Take a deep breath and think carefully.
0.08 - You are highly intelligent. Answer the following math question. I really need your help!
0.08 - You are an expert mathematician. None. None.
0.08 - You are an expert mathematician. None. This will be fun!
0.08 - You are an expert mathematician. None. Take a deep breath and think carefully.
0.08 - You are an expert mathematician. None. I really need your help!
0.08 - You are an expert mathematician. Solve the following math problem. None.
0.08 - You are an expert mathematician. Solve the following math problem. This will be fun!
0.08 - You are an expert mathematician. Solve the following math problem. Take a deep breath and think carefully.
0.08 - You are an expert mathematician. Solve the following math problem. I really need your help!
0.08 - You are an expert mathematician. Answer the following math question. None.
0.08 - You are an expert mathematician. Answer the following math question. This will be fun!
0.08 - You are an expert mathematician. Answer the following math question. Take a deep breath and think carefully.
0.08 - You are an expert mathematician. Answer the following math question. I really need your help!
0.08 - You are a professor of mathematics. None. None.
0.08 - You are a professor of mathematics. None. This will be fun!
0.08 - You are a professor of mathematics. None. Take a deep breath and think carefully.
0.08 - You are a professor of mathematics. None. I really need your help!
0.08 - You are a professor of mathematics. Solve the following math problem. None.
0.08 - You are a professor of mathematics. Solve the following math problem. This will be fun!
0.08 - You are a professor of mathematics. Solve the following math problem. Take a deep breath and think carefully.
0.08 - You are a professor of mathematics. Solve the following math problem. I really need your help!
0.08 - You are a professor of mathematics. Answer the following math question. None.
0.08 - You are a professor of mathematics. Answer the following math question. This will be fun!
0.08 - You are a professor of mathematics. Answer the following math question. Take a deep breath and think carefully.
0.08 - You are a professor of mathematics. Answer the following math question. I really need your help!
\end{verbatim}

\subsection{Mistral-7B CoT=No NoQ=50}
\label{results-mistral-cot=no-q=50}
 
EM - Prompt
 
\begin{verbatim}
0.1 - You are a professor of mathematics. Solve the following math problem. Take a deep breath and think carefully.
0.12 - None. None. None.
0.12 - None. None. This will be fun!
0.12 - None. None. Take a deep breath and think carefully.
0.12 - None. None. I really need your help!
0.12 - None. Solve the following math problem. None.
0.12 - None. Solve the following math problem. This will be fun!
0.12 - None. Solve the following math problem. Take a deep breath and think carefully.
0.12 - None. Solve the following math problem. I really need your help!
0.12 - None. Answer the following math question. None.
0.12 - None. Answer the following math question. This will be fun!
0.12 - None. Answer the following math question. Take a deep breath and think carefully.
0.12 - None. Answer the following math question. I really need your help!
0.12 - You are as smart as ChatGPT. None. None.
0.12 - You are as smart as ChatGPT. None. This will be fun!
0.12 - You are as smart as ChatGPT. None. Take a deep breath and think carefully.
0.12 - You are as smart as ChatGPT. None. I really need your help!
0.12 - You are as smart as ChatGPT. Solve the following math problem. None.
0.12 - You are as smart as ChatGPT. Solve the following math problem. This will be fun!
0.12 - You are as smart as ChatGPT. Solve the following math problem. Take a deep breath and think carefully.
0.12 - You are as smart as ChatGPT. Solve the following math problem. I really need your help!
0.12 - You are as smart as ChatGPT. Answer the following math question. None.
0.12 - You are as smart as ChatGPT. Answer the following math question. This will be fun!
0.12 - You are as smart as ChatGPT. Answer the following math question. Take a deep breath and think carefully.
0.12 - You are as smart as ChatGPT. Answer the following math question. I really need your help!
0.12 - You are highly intelligent. None. None.
0.12 - You are highly intelligent. None. This will be fun!
0.12 - You are highly intelligent. None. Take a deep breath and think carefully.
0.12 - You are highly intelligent. None. I really need your help!
0.12 - You are highly intelligent. Solve the following math problem. None.
0.12 - You are highly intelligent. Solve the following math problem. This will be fun!
0.12 - You are highly intelligent. Solve the following math problem. Take a deep breath and think carefully.
0.12 - You are highly intelligent. Solve the following math problem. I really need your help!
0.12 - You are highly intelligent. Answer the following math question. None.
0.12 - You are highly intelligent. Answer the following math question. This will be fun!
0.12 - You are highly intelligent. Answer the following math question. Take a deep breath and think carefully.
0.12 - You are highly intelligent. Answer the following math question. I really need your help!
0.12 - You are an expert mathematician. None. None.
0.12 - You are an expert mathematician. None. This will be fun!
0.12 - You are an expert mathematician. None. Take a deep breath and think carefully.
0.12 - You are an expert mathematician. None. I really need your help!
0.12 - You are an expert mathematician. Solve the following math problem. None.
0.12 - You are an expert mathematician. Solve the following math problem. This will be fun!
0.12 - You are an expert mathematician. Solve the following math problem. Take a deep breath and think carefully.
0.12 - You are an expert mathematician. Solve the following math problem. I really need your help!
0.12 - You are an expert mathematician. Answer the following math question. None.
0.12 - You are an expert mathematician. Answer the following math question. This will be fun!
0.12 - You are an expert mathematician. Answer the following math question. Take a deep breath and think carefully.
0.12 - You are an expert mathematician. Answer the following math question. I really need your help!
0.12 - You are a professor of mathematics. None. None.
0.12 - You are a professor of mathematics. None. This will be fun!
0.12 - You are a professor of mathematics. None. Take a deep breath and think carefully.
0.12 - You are a professor of mathematics. None. I really need your help!
0.12 - You are a professor of mathematics. Solve the following math problem. None.
0.12 - You are a professor of mathematics. Solve the following math problem. This will be fun!
0.12 - You are a professor of mathematics. Solve the following math problem. I really need your help!
0.12 - You are a professor of mathematics. Answer the following math question. None.
0.12 - You are a professor of mathematics. Answer the following math question. This will be fun!
0.12 - You are a professor of mathematics. Answer the following math question. Take a deep breath and think carefully.
0.12 - You are a professor of mathematics. Answer the following math question. I really need your help!
\end{verbatim}

\subsection{Mistral-7B CoT=No NoQ=100}
\label{results-mistral-cot=no-q=100}
 
EM - Prompt
 
\begin{verbatim}
0.08 - None. None. Take a deep breath and think carefully.
0.09 - None. None. None.
0.09 - You are highly intelligent. None. None.
0.09 - You are highly intelligent. None. This will be fun!
0.09 - You are highly intelligent. None. Take a deep breath and think carefully.
0.09 - You are highly intelligent. None. I really need your help!
0.1 - None. None. This will be fun!
0.1 - None. None. I really need your help!
0.1 - None. Solve the following math problem. None.
0.1 - None. Solve the following math problem. I really need your help!
0.1 - None. Answer the following math question. None.
0.1 - None. Answer the following math question. This will be fun!
0.1 - None. Answer the following math question. Take a deep breath and think carefully.
0.1 - None. Answer the following math question. I really need your help!
0.1 - You are as smart as ChatGPT. None. None.
0.1 - You are as smart as ChatGPT. None. This will be fun!
0.1 - You are as smart as ChatGPT. None. Take a deep breath and think carefully.
0.1 - You are as smart as ChatGPT. None. I really need your help!
0.1 - You are highly intelligent. Solve the following math problem. This will be fun!
0.1 - You are highly intelligent. Answer the following math question. This will be fun!
0.1 - You are a professor of mathematics. Solve the following math problem. Take a deep breath and think carefully.
0.11 - None. Solve the following math problem. This will be fun!
0.11 - None. Solve the following math problem. Take a deep breath and think carefully.
0.11 - You are as smart as ChatGPT. Solve the following math problem. None.
0.11 - You are as smart as ChatGPT. Solve the following math problem. This will be fun!
0.11 - You are as smart as ChatGPT. Solve the following math problem. Take a deep breath and think carefully.
0.11 - You are as smart as ChatGPT. Solve the following math problem. I really need your help!
0.11 - You are as smart as ChatGPT. Answer the following math question. None.
0.11 - You are as smart as ChatGPT. Answer the following math question. This will be fun!
0.11 - You are as smart as ChatGPT. Answer the following math question. Take a deep breath and think carefully.
0.11 - You are as smart as ChatGPT. Answer the following math question. I really need your help!
0.11 - You are highly intelligent. Solve the following math problem. None.
0.11 - You are highly intelligent. Solve the following math problem. Take a deep breath and think carefully.
0.11 - You are highly intelligent. Solve the following math problem. I really need your help!
0.11 - You are highly intelligent. Answer the following math question. None.
0.11 - You are highly intelligent. Answer the following math question. Take a deep breath and think carefully.
0.11 - You are highly intelligent. Answer the following math question. I really need your help!
0.11 - You are an expert mathematician. None. None.
0.11 - You are an expert mathematician. None. This will be fun!
0.11 - You are an expert mathematician. None. Take a deep breath and think carefully.
0.11 - You are an expert mathematician. None. I really need your help!
0.11 - You are an expert mathematician. Solve the following math problem. None.
0.11 - You are an expert mathematician. Solve the following math problem. This will be fun!
0.11 - You are an expert mathematician. Solve the following math problem. Take a deep breath and think carefully.
0.11 - You are an expert mathematician. Solve the following math problem. I really need your help!
0.11 - You are an expert mathematician. Answer the following math question. None.
0.11 - You are an expert mathematician. Answer the following math question. This will be fun!
0.11 - You are an expert mathematician. Answer the following math question. Take a deep breath and think carefully.
0.11 - You are an expert mathematician. Answer the following math question. I really need your help!
0.11 - You are a professor of mathematics. None. None.
0.11 - You are a professor of mathematics. None. This will be fun!
0.11 - You are a professor of mathematics. None. Take a deep breath and think carefully.
0.11 - You are a professor of mathematics. None. I really need your help!
0.11 - You are a professor of mathematics. Solve the following math problem. None.
0.11 - You are a professor of mathematics. Solve the following math problem. This will be fun!
0.11 - You are a professor of mathematics. Solve the following math problem. I really need your help!
0.11 - You are a professor of mathematics. Answer the following math question. None.
0.11 - You are a professor of mathematics. Answer the following math question. This will be fun!
0.11 - You are a professor of mathematics. Answer the following math question. Take a deep breath and think carefully.
0.11 - You are a professor of mathematics. Answer the following math question. I really need your help!
\end{verbatim}

\subsection{Mistral-7B CoT=Yes NoQ=10}
\label{results-mistral-cot=yes-q=10}
 
EM - Prompt
 
\begin{verbatim}
0.2 - None. None. None.
0.2 - You are highly intelligent. Solve the following math problem. This will be fun!
0.3 - None. Answer the following math question. None.
0.3 - None. Answer the following math question. This will be fun!
0.3 - None. Answer the following math question. Take a deep breath and think carefully.
0.3 - You are as smart as ChatGPT. Answer the following math question. I really need your help!
0.3 - You are highly intelligent. Solve the following math problem. Take a deep breath and think carefully.
0.3 - You are highly intelligent. Answer the following math question. None.
0.3 - You are highly intelligent. Answer the following math question. This will be fun!
0.3 - You are highly intelligent. Answer the following math question. I really need your help!
0.3 - You are an expert mathematician. Solve the following math problem. This will be fun!
0.3 - You are an expert mathematician. Solve the following math problem. Take a deep breath and think carefully.
0.3 - You are an expert mathematician. Answer the following math question. This will be fun!
0.3 - You are an expert mathematician. Answer the following math question. I really need your help!
0.3 - You are a professor of mathematics. Solve the following math problem. This will be fun!
0.3 - You are a professor of mathematics. Solve the following math problem. Take a deep breath and think carefully.
0.4 - None. None. Take a deep breath and think carefully.
0.4 - None. None. I really need your help!
0.4 - None. Solve the following math problem. None.
0.4 - None. Solve the following math problem. This will be fun!
0.4 - None. Solve the following math problem. Take a deep breath and think carefully.
0.4 - None. Solve the following math problem. I really need your help!
0.4 - You are as smart as ChatGPT. None. This will be fun!
0.4 - You are as smart as ChatGPT. None. Take a deep breath and think carefully.
0.4 - You are as smart as ChatGPT. None. I really need your help!
0.4 - You are as smart as ChatGPT. Solve the following math problem. None.
0.4 - You are as smart as ChatGPT. Solve the following math problem. I really need your help!
0.4 - You are as smart as ChatGPT. Answer the following math question. This will be fun!
0.4 - You are as smart as ChatGPT. Answer the following math question. Take a deep breath and think carefully.
0.4 - You are highly intelligent. None. None.
0.4 - You are highly intelligent. None. This will be fun!
0.4 - You are highly intelligent. None. Take a deep breath and think carefully.
0.4 - You are highly intelligent. None. I really need your help!
0.4 - You are highly intelligent. Solve the following math problem. None.
0.4 - You are highly intelligent. Solve the following math problem. I really need your help!
0.4 - You are highly intelligent. Answer the following math question. Take a deep breath and think carefully.
0.4 - You are an expert mathematician. None. None.
0.4 - You are an expert mathematician. None. This will be fun!
0.4 - You are an expert mathematician. None. Take a deep breath and think carefully.
0.4 - You are an expert mathematician. None. I really need your help!
0.4 - You are an expert mathematician. Solve the following math problem. None.
0.4 - You are an expert mathematician. Solve the following math problem. I really need your help!
0.4 - You are an expert mathematician. Answer the following math question. None.
0.4 - You are an expert mathematician. Answer the following math question. Take a deep breath and think carefully.
0.4 - You are a professor of mathematics. None. None.
0.4 - You are a professor of mathematics. None. This will be fun!
0.4 - You are a professor of mathematics. None. Take a deep breath and think carefully.
0.4 - You are a professor of mathematics. None. I really need your help!
0.4 - You are a professor of mathematics. Solve the following math problem. None.
0.4 - You are a professor of mathematics. Solve the following math problem. I really need your help!
0.4 - You are a professor of mathematics. Answer the following math question. None.
0.4 - You are a professor of mathematics. Answer the following math question. This will be fun!
0.4 - You are a professor of mathematics. Answer the following math question. Take a deep breath and think carefully.
0.4 - You are a professor of mathematics. Answer the following math question. I really need your help!
0.5 - None. None. This will be fun!
0.5 - None. Answer the following math question. I really need your help!
0.5 - You are as smart as ChatGPT. None. None.
0.5 - You are as smart as ChatGPT. Solve the following math problem. This will be fun!
0.5 - You are as smart as ChatGPT. Solve the following math problem. Take a deep breath and think carefully.
0.5 - You are as smart as ChatGPT. Answer the following math question. None.
\end{verbatim}

\subsection{Mistral-7B CoT=Yes NoQ=25}
\label{results-mistral-cot=yes-q=25}
 
EM - Prompt
 
\begin{verbatim}
0.28 - None. None. None.
0.28 - You are highly intelligent. Solve the following math problem. This will be fun!
0.28 - You are highly intelligent. Answer the following math question. This will be fun!
0.28 - You are highly intelligent. Answer the following math question. I really need your help!
0.32 - None. None. I really need your help!
0.32 - None. Answer the following math question. This will be fun!
0.32 - None. Answer the following math question. Take a deep breath and think carefully.
0.32 - You are as smart as ChatGPT. None. This will be fun!
0.32 - You are as smart as ChatGPT. None. Take a deep breath and think carefully.
0.32 - You are as smart as ChatGPT. None. I really need your help!
0.32 - You are as smart as ChatGPT. Solve the following math problem. None.
0.32 - You are as smart as ChatGPT. Answer the following math question. Take a deep breath and think carefully.
0.32 - You are as smart as ChatGPT. Answer the following math question. I really need your help!
0.32 - You are highly intelligent. None. This will be fun!
0.32 - You are highly intelligent. None. I really need your help!
0.32 - You are highly intelligent. Solve the following math problem. Take a deep breath and think carefully.
0.32 - You are highly intelligent. Answer the following math question. Take a deep breath and think carefully.
0.32 - You are a professor of mathematics. Solve the following math problem. Take a deep breath and think carefully.
0.32 - You are a professor of mathematics. Answer the following math question. Take a deep breath and think carefully.
0.36 - None. None. Take a deep breath and think carefully.
0.36 - None. Solve the following math problem. This will be fun!
0.36 - None. Solve the following math problem. I really need your help!
0.36 - None. Answer the following math question. None.
0.36 - You are as smart as ChatGPT. Solve the following math problem. Take a deep breath and think carefully.
0.36 - You are highly intelligent. None. None.
0.36 - You are highly intelligent. None. Take a deep breath and think carefully.
0.36 - You are highly intelligent. Solve the following math problem. None.
0.36 - You are highly intelligent. Solve the following math problem. I really need your help!
0.36 - You are highly intelligent. Answer the following math question. None.
0.36 - You are an expert mathematician. None. Take a deep breath and think carefully.
0.36 - You are an expert mathematician. Solve the following math problem. Take a deep breath and think carefully.
0.36 - You are a professor of mathematics. None. Take a deep breath and think carefully.
0.36 - You are a professor of mathematics. None. I really need your help!
0.36 - You are a professor of mathematics. Solve the following math problem. This will be fun!
0.4 - None. None. This will be fun!
0.4 - None. Solve the following math problem. None.
0.4 - None. Solve the following math problem. Take a deep breath and think carefully.
0.4 - You are as smart as ChatGPT. None. None.
0.4 - You are as smart as ChatGPT. Solve the following math problem. This will be fun!
0.4 - You are as smart as ChatGPT. Solve the following math problem. I really need your help!
0.4 - You are as smart as ChatGPT. Answer the following math question. None.
0.4 - You are as smart as ChatGPT. Answer the following math question. This will be fun!
0.4 - You are an expert mathematician. None. None.
0.4 - You are an expert mathematician. None. I really need your help!
0.4 - You are an expert mathematician. Solve the following math problem. None.
0.4 - You are an expert mathematician. Solve the following math problem. This will be fun!
0.4 - You are an expert mathematician. Answer the following math question. None.
0.4 - You are an expert mathematician. Answer the following math question. This will be fun!
0.4 - You are an expert mathematician. Answer the following math question. I really need your help!
0.4 - You are a professor of mathematics. None. This will be fun!
0.4 - You are a professor of mathematics. Solve the following math problem. None.
0.4 - You are a professor of mathematics. Solve the following math problem. I really need your help!
0.4 - You are a professor of mathematics. Answer the following math question. None.
0.4 - You are a professor of mathematics. Answer the following math question. This will be fun!
0.4 - You are a professor of mathematics. Answer the following math question. I really need your help!
0.44 - None. Answer the following math question. I really need your help!
0.44 - You are an expert mathematician. None. This will be fun!
0.44 - You are an expert mathematician. Solve the following math problem. I really need your help!
0.44 - You are a professor of mathematics. None. None.
0.48 - You are an expert mathematician. Answer the following math question. Take a deep breath and think carefully.
\end{verbatim}

\subsection{Mistral-7B CoT=Yes NoQ=50}
\label{results-mistral-cot=yes-q=50}
 
EM - Prompt
 
\begin{verbatim}
0.32 - None. None. None.
0.34 - None. Answer the following math question. None.
0.34 - You are a professor of mathematics. Answer the following math question. Take a deep breath and think carefully.
0.36 - None. None. I really need your help!
0.36 - You are as smart as ChatGPT. Answer the following math question. Take a deep breath and think carefully.
0.36 - You are as smart as ChatGPT. Answer the following math question. I really need your help!
0.36 - You are highly intelligent. Answer the following math question. This will be fun!
0.36 - You are highly intelligent. Answer the following math question. I really need your help!
0.36 - You are an expert mathematician. Solve the following math problem. Take a deep breath and think carefully.
0.36 - You are a professor of mathematics. None. Take a deep breath and think carefully.
0.36 - You are a professor of mathematics. Solve the following math problem. None.
0.36 - You are a professor of mathematics. Solve the following math problem. Take a deep breath and think carefully.
0.38 - None. Answer the following math question. This will be fun!
0.38 - None. Answer the following math question. Take a deep breath and think carefully.
0.38 - You are as smart as ChatGPT. None. This will be fun!
0.38 - You are as smart as ChatGPT. None. I really need your help!
0.38 - You are as smart as ChatGPT. Solve the following math problem. None.
0.38 - You are highly intelligent. None. This will be fun!
0.38 - You are highly intelligent. None. I really need your help!
0.38 - You are highly intelligent. Solve the following math problem. This will be fun!
0.38 - You are highly intelligent. Solve the following math problem. Take a deep breath and think carefully.
0.38 - You are highly intelligent. Solve the following math problem. I really need your help!
0.38 - You are highly intelligent. Answer the following math question. None.
0.38 - You are highly intelligent. Answer the following math question. Take a deep breath and think carefully.
0.38 - You are an expert mathematician. Solve the following math problem. None.
0.38 - You are an expert mathematician. Solve the following math problem. This will be fun!
0.38 - You are an expert mathematician. Answer the following math question. None.
0.38 - You are an expert mathematician. Answer the following math question. This will be fun!
0.38 - You are a professor of mathematics. None. I really need your help!
0.38 - You are a professor of mathematics. Solve the following math problem. This will be fun!
0.38 - You are a professor of mathematics. Solve the following math problem. I really need your help!
0.38 - You are a professor of mathematics. Answer the following math question. None.
0.4 - None. Solve the following math problem. None.
0.4 - None. Solve the following math problem. This will be fun!
0.4 - None. Solve the following math problem. Take a deep breath and think carefully.
0.4 - None. Solve the following math problem. I really need your help!
0.4 - None. Answer the following math question. I really need your help!
0.4 - You are as smart as ChatGPT. None. None.
0.4 - You are as smart as ChatGPT. None. Take a deep breath and think carefully.
0.4 - You are as smart as ChatGPT. Solve the following math problem. Take a deep breath and think carefully.
0.4 - You are highly intelligent. None. Take a deep breath and think carefully.
0.4 - You are highly intelligent. Solve the following math problem. None.
0.4 - You are an expert mathematician. None. None.
0.4 - You are an expert mathematician. None. Take a deep breath and think carefully.
0.4 - You are an expert mathematician. None. I really need your help!
0.4 - You are an expert mathematician. Solve the following math problem. I really need your help!
0.4 - You are an expert mathematician. Answer the following math question. I really need your help!
0.4 - You are a professor of mathematics. None. This will be fun!
0.4 - You are a professor of mathematics. Answer the following math question. I really need your help!
0.42 - None. None. This will be fun!
0.42 - None. None. Take a deep breath and think carefully.
0.42 - You are as smart as ChatGPT. Solve the following math problem. This will be fun!
0.42 - You are as smart as ChatGPT. Solve the following math problem. I really need your help!
0.42 - You are highly intelligent. None. None.
0.42 - You are an expert mathematician. None. This will be fun!
0.42 - You are a professor of mathematics. Answer the following math question. This will be fun!
0.44 - You are as smart as ChatGPT. Answer the following math question. None.
0.44 - You are as smart as ChatGPT. Answer the following math question. This will be fun!
0.44 - You are an expert mathematician. Answer the following math question. Take a deep breath and think carefully.
0.44 - You are a professor of mathematics. None. None.
\end{verbatim}

\subsection{Mistral-7B CoT=Yes NoQ=100}
\label{results-mistral-cot=yes-q=100}
 
EM - Prompt
 
\begin{verbatim}
0.35 - None. None. None.
0.37 - None. None. I really need your help!
0.37 - You are highly intelligent. Answer the following math question. This will be fun!
0.37 - You are a professor of mathematics. Answer the following math question. Take a deep breath and think carefully.
0.38 - You are as smart as ChatGPT. None. This will be fun!
0.38 - You are as smart as ChatGPT. Answer the following math question. I really need your help!
0.38 - You are highly intelligent. None. I really need your help!
0.38 - You are an expert mathematician. Solve the following math problem. Take a deep breath and think carefully.
0.39 - None. None. Take a deep breath and think carefully.
0.39 - None. Answer the following math question. None.
0.39 - None. Answer the following math question. I really need your help!
0.39 - You are as smart as ChatGPT. None. I really need your help!
0.39 - You are as smart as ChatGPT. Answer the following math question. Take a deep breath and think carefully.
0.39 - You are highly intelligent. Solve the following math problem. I really need your help!
0.39 - You are highly intelligent. Answer the following math question. None.
0.39 - You are an expert mathematician. Solve the following math problem. This will be fun!
0.39 - You are an expert mathematician. Answer the following math question. This will be fun!
0.39 - You are a professor of mathematics. Solve the following math problem. Take a deep breath and think carefully.
0.39 - You are a professor of mathematics. Answer the following math question. None.
0.4 - None. Solve the following math problem. This will be fun!
0.4 - None. Solve the following math problem. I really need your help!
0.4 - You are as smart as ChatGPT. None. None.
0.4 - You are highly intelligent. Solve the following math problem. This will be fun!
0.4 - You are highly intelligent. Solve the following math problem. Take a deep breath and think carefully.
0.4 - You are highly intelligent. Answer the following math question. Take a deep breath and think carefully.
0.4 - You are an expert mathematician. Solve the following math problem. None.
0.4 - You are an expert mathematician. Solve the following math problem. I really need your help!
0.4 - You are an expert mathematician. Answer the following math question. None.
0.4 - You are an expert mathematician. Answer the following math question. I really need your help!
0.4 - You are a professor of mathematics. None. This will be fun!
0.4 - You are a professor of mathematics. None. Take a deep breath and think carefully.
0.4 - You are a professor of mathematics. None. I really need your help!
0.4 - You are a professor of mathematics. Solve the following math problem. None.
0.4 - You are a professor of mathematics. Solve the following math problem. This will be fun!
0.4 - You are a professor of mathematics. Solve the following math problem. I really need your help!
0.41 - None. Solve the following math problem. None.
0.41 - You are as smart as ChatGPT. None. Take a deep breath and think carefully.
0.41 - You are as smart as ChatGPT. Solve the following math problem. None.
0.41 - You are highly intelligent. None. This will be fun!
0.41 - You are highly intelligent. None. Take a deep breath and think carefully.
0.41 - You are highly intelligent. Answer the following math question. I really need your help!
0.41 - You are an expert mathematician. None. I really need your help!
0.41 - You are a professor of mathematics. Answer the following math question. I really need your help!
0.42 - None. Solve the following math problem. Take a deep breath and think carefully.
0.42 - You are as smart as ChatGPT. Solve the following math problem. This will be fun!
0.42 - You are as smart as ChatGPT. Solve the following math problem. I really need your help!
0.42 - You are highly intelligent. None. None.
0.42 - You are highly intelligent. Solve the following math problem. None.
0.42 - You are an expert mathematician. None. None.
0.42 - You are an expert mathematician. None. This will be fun!
0.42 - You are an expert mathematician. None. Take a deep breath and think carefully.
0.42 - You are a professor of mathematics. Answer the following math question. This will be fun!
0.43 - None. None. This will be fun!
0.43 - None. Answer the following math question. This will be fun!
0.43 - None. Answer the following math question. Take a deep breath and think carefully.
0.43 - You are as smart as ChatGPT. Answer the following math question. None.
0.43 - You are as smart as ChatGPT. Answer the following math question. This will be fun!
0.43 - You are an expert mathematician. Answer the following math question. Take a deep breath and think carefully.
0.43 - You are a professor of mathematics. None. None.
0.44 - You are as smart as ChatGPT. Solve the following math problem. Take a deep breath and think carefully.
\end{verbatim}

\subsection{Llama2-13B CoT=No NoQ=10}
\label{results-llama2-13-cot=no-q=10}
 
EM - Prompt
 
\begin{verbatim}
0.1 - None. None. None.
0.1 - None. None. This will be fun!
0.1 - None. None. Take a deep breath and think carefully.
0.1 - None. None. I really need your help!
0.1 - None. Solve the following math problem. None.
0.1 - None. Solve the following math problem. This will be fun!
0.1 - None. Solve the following math problem. Take a deep breath and think carefully.
0.1 - None. Solve the following math problem. I really need your help!
0.1 - None. Answer the following math question. None.
0.1 - None. Answer the following math question. This will be fun!
0.1 - None. Answer the following math question. Take a deep breath and think carefully.
0.1 - None. Answer the following math question. I really need your help!
0.1 - You are as smart as ChatGPT. None. None.
0.1 - You are as smart as ChatGPT. None. This will be fun!
0.1 - You are as smart as ChatGPT. None. Take a deep breath and think carefully.
0.1 - You are as smart as ChatGPT. None. I really need your help!
0.1 - You are as smart as ChatGPT. Solve the following math problem. None.
0.1 - You are as smart as ChatGPT. Solve the following math problem. This will be fun!
0.1 - You are as smart as ChatGPT. Solve the following math problem. Take a deep breath and think carefully.
0.1 - You are as smart as ChatGPT. Solve the following math problem. I really need your help!
0.1 - You are as smart as ChatGPT. Answer the following math question. None.
0.1 - You are as smart as ChatGPT. Answer the following math question. This will be fun!
0.1 - You are as smart as ChatGPT. Answer the following math question. Take a deep breath and think carefully.
0.1 - You are as smart as ChatGPT. Answer the following math question. I really need your help!
0.1 - You are highly intelligent. None. None.
0.1 - You are highly intelligent. None. This will be fun!
0.1 - You are highly intelligent. None. Take a deep breath and think carefully.
0.1 - You are highly intelligent. None. I really need your help!
0.1 - You are highly intelligent. Solve the following math problem. None.
0.1 - You are highly intelligent. Solve the following math problem. This will be fun!
0.1 - You are highly intelligent. Solve the following math problem. Take a deep breath and think carefully.
0.1 - You are highly intelligent. Solve the following math problem. I really need your help!
0.1 - You are highly intelligent. Answer the following math question. None.
0.1 - You are highly intelligent. Answer the following math question. This will be fun!
0.1 - You are highly intelligent. Answer the following math question. Take a deep breath and think carefully.
0.1 - You are highly intelligent. Answer the following math question. I really need your help!
0.1 - You are an expert mathematician. None. None.
0.1 - You are an expert mathematician. None. This will be fun!
0.1 - You are an expert mathematician. None. Take a deep breath and think carefully.
0.1 - You are an expert mathematician. None. I really need your help!
0.1 - You are an expert mathematician. Solve the following math problem. None.
0.1 - You are an expert mathematician. Solve the following math problem. This will be fun!
0.1 - You are an expert mathematician. Solve the following math problem. Take a deep breath and think carefully.
0.1 - You are an expert mathematician. Solve the following math problem. I really need your help!
0.1 - You are an expert mathematician. Answer the following math question. None.
0.1 - You are an expert mathematician. Answer the following math question. This will be fun!
0.1 - You are an expert mathematician. Answer the following math question. Take a deep breath and think carefully.
0.1 - You are an expert mathematician. Answer the following math question. I really need your help!
0.1 - You are a professor of mathematics. None. None.
0.1 - You are a professor of mathematics. None. This will be fun!
0.1 - You are a professor of mathematics. None. Take a deep breath and think carefully.
0.1 - You are a professor of mathematics. None. I really need your help!
0.1 - You are a professor of mathematics. Solve the following math problem. None.
0.1 - You are a professor of mathematics. Solve the following math problem. This will be fun!
0.1 - You are a professor of mathematics. Solve the following math problem. Take a deep breath and think carefully.
0.1 - You are a professor of mathematics. Solve the following math problem. I really need your help!
0.1 - You are a professor of mathematics. Answer the following math question. None.
0.1 - You are a professor of mathematics. Answer the following math question. This will be fun!
0.1 - You are a professor of mathematics. Answer the following math question. Take a deep breath and think carefully.
0.1 - You are a professor of mathematics. Answer the following math question. I really need your help!
\end{verbatim}

\subsection{Llama2-13B CoT=No NoQ=25}
\label{results-llama2-13-cot=no-q=25}
 
EM - Prompt
 
\begin{verbatim}
0.08 - None. None. None.
0.08 - None. None. This will be fun!
0.08 - None. None. Take a deep breath and think carefully.
0.08 - None. None. I really need your help!
0.08 - None. Solve the following math problem. None.
0.08 - None. Solve the following math problem. This will be fun!
0.08 - None. Solve the following math problem. Take a deep breath and think carefully.
0.08 - None. Solve the following math problem. I really need your help!
0.08 - None. Answer the following math question. None.
0.08 - None. Answer the following math question. This will be fun!
0.08 - None. Answer the following math question. Take a deep breath and think carefully.
0.08 - None. Answer the following math question. I really need your help!
0.08 - You are as smart as ChatGPT. None. None.
0.08 - You are as smart as ChatGPT. None. This will be fun!
0.08 - You are as smart as ChatGPT. None. Take a deep breath and think carefully.
0.08 - You are as smart as ChatGPT. None. I really need your help!
0.08 - You are as smart as ChatGPT. Solve the following math problem. None.
0.08 - You are as smart as ChatGPT. Solve the following math problem. This will be fun!
0.08 - You are as smart as ChatGPT. Solve the following math problem. Take a deep breath and think carefully.
0.08 - You are as smart as ChatGPT. Answer the following math question. None.
0.08 - You are as smart as ChatGPT. Answer the following math question. This will be fun!
0.08 - You are as smart as ChatGPT. Answer the following math question. Take a deep breath and think carefully.
0.08 - You are as smart as ChatGPT. Answer the following math question. I really need your help!
0.08 - You are highly intelligent. None. None.
0.08 - You are highly intelligent. None. This will be fun!
0.08 - You are highly intelligent. None. Take a deep breath and think carefully.
0.08 - You are highly intelligent. None. I really need your help!
0.08 - You are highly intelligent. Solve the following math problem. None.
0.08 - You are highly intelligent. Solve the following math problem. This will be fun!
0.08 - You are highly intelligent. Solve the following math problem. Take a deep breath and think carefully.
0.08 - You are highly intelligent. Answer the following math question. None.
0.08 - You are highly intelligent. Answer the following math question. This will be fun!
0.08 - You are highly intelligent. Answer the following math question. Take a deep breath and think carefully.
0.08 - You are an expert mathematician. None. None.
0.08 - You are an expert mathematician. None. This will be fun!
0.08 - You are an expert mathematician. None. Take a deep breath and think carefully.
0.08 - You are an expert mathematician. Solve the following math problem. None.
0.08 - You are an expert mathematician. Solve the following math problem. This will be fun!
0.08 - You are an expert mathematician. Solve the following math problem. Take a deep breath and think carefully.
0.08 - You are an expert mathematician. Answer the following math question. None.
0.08 - You are an expert mathematician. Answer the following math question. This will be fun!
0.08 - You are an expert mathematician. Answer the following math question. Take a deep breath and think carefully.
0.08 - You are a professor of mathematics. None. None.
0.08 - You are a professor of mathematics. None. This will be fun!
0.08 - You are a professor of mathematics. None. Take a deep breath and think carefully.
0.08 - You are a professor of mathematics. None. I really need your help!
0.08 - You are a professor of mathematics. Solve the following math problem. None.
0.08 - You are a professor of mathematics. Solve the following math problem. This will be fun!
0.08 - You are a professor of mathematics. Solve the following math problem. Take a deep breath and think carefully.
0.08 - You are a professor of mathematics. Answer the following math question. None.
0.08 - You are a professor of mathematics. Answer the following math question. This will be fun!
0.08 - You are a professor of mathematics. Answer the following math question. Take a deep breath and think carefully.
0.12 - You are as smart as ChatGPT. Solve the following math problem. I really need your help!
0.12 - You are highly intelligent. Solve the following math problem. I really need your help!
0.12 - You are highly intelligent. Answer the following math question. I really need your help!
0.12 - You are an expert mathematician. None. I really need your help!
0.12 - You are an expert mathematician. Solve the following math problem. I really need your help!
0.12 - You are an expert mathematician. Answer the following math question. I really need your help!
0.12 - You are a professor of mathematics. Solve the following math problem. I really need your help!
0.12 - You are a professor of mathematics. Answer the following math question. I really need your help!
\end{verbatim}

\subsection{Llama2-13B CoT=No NoQ=50}
\label{results-llama2-13-cot=no-q=50}
 
EM - Prompt
 
\begin{verbatim}
0.08 - None. None. None.
0.08 - None. None. This will be fun!
0.08 - None. None. Take a deep breath and think carefully.
0.08 - None. None. I really need your help!
0.08 - None. Solve the following math problem. None.
0.08 - None. Solve the following math problem. This will be fun!
0.08 - None. Solve the following math problem. Take a deep breath and think carefully.
0.08 - None. Solve the following math problem. I really need your help!
0.08 - None. Answer the following math question. None.
0.08 - None. Answer the following math question. This will be fun!
0.08 - None. Answer the following math question. Take a deep breath and think carefully.
0.08 - None. Answer the following math question. I really need your help!
0.08 - You are as smart as ChatGPT. None. None.
0.08 - You are as smart as ChatGPT. None. This will be fun!
0.08 - You are as smart as ChatGPT. None. Take a deep breath and think carefully.
0.08 - You are as smart as ChatGPT. None. I really need your help!
0.08 - You are as smart as ChatGPT. Solve the following math problem. None.
0.08 - You are as smart as ChatGPT. Solve the following math problem. This will be fun!
0.08 - You are as smart as ChatGPT. Solve the following math problem. Take a deep breath and think carefully.
0.08 - You are as smart as ChatGPT. Answer the following math question. None.
0.08 - You are as smart as ChatGPT. Answer the following math question. This will be fun!
0.08 - You are as smart as ChatGPT. Answer the following math question. Take a deep breath and think carefully.
0.08 - You are as smart as ChatGPT. Answer the following math question. I really need your help!
0.08 - You are highly intelligent. None. None.
0.08 - You are highly intelligent. None. This will be fun!
0.08 - You are highly intelligent. None. Take a deep breath and think carefully.
0.08 - You are highly intelligent. None. I really need your help!
0.08 - You are highly intelligent. Solve the following math problem. None.
0.08 - You are highly intelligent. Solve the following math problem. This will be fun!
0.08 - You are highly intelligent. Solve the following math problem. Take a deep breath and think carefully.
0.08 - You are highly intelligent. Answer the following math question. None.
0.08 - You are highly intelligent. Answer the following math question. This will be fun!
0.08 - You are highly intelligent. Answer the following math question. Take a deep breath and think carefully.
0.08 - You are an expert mathematician. None. None.
0.08 - You are an expert mathematician. None. This will be fun!
0.08 - You are an expert mathematician. None. Take a deep breath and think carefully.
0.08 - You are an expert mathematician. Solve the following math problem. None.
0.08 - You are an expert mathematician. Solve the following math problem. This will be fun!
0.08 - You are an expert mathematician. Solve the following math problem. Take a deep breath and think carefully.
0.08 - You are an expert mathematician. Answer the following math question. None.
0.08 - You are an expert mathematician. Answer the following math question. This will be fun!
0.08 - You are an expert mathematician. Answer the following math question. Take a deep breath and think carefully.
0.08 - You are a professor of mathematics. None. None.
0.08 - You are a professor of mathematics. None. This will be fun!
0.08 - You are a professor of mathematics. None. Take a deep breath and think carefully.
0.08 - You are a professor of mathematics. None. I really need your help!
0.08 - You are a professor of mathematics. Solve the following math problem. None.
0.08 - You are a professor of mathematics. Solve the following math problem. This will be fun!
0.08 - You are a professor of mathematics. Solve the following math problem. Take a deep breath and think carefully.
0.08 - You are a professor of mathematics. Answer the following math question. None.
0.08 - You are a professor of mathematics. Answer the following math question. This will be fun!
0.08 - You are a professor of mathematics. Answer the following math question. Take a deep breath and think carefully.
0.1 - You are as smart as ChatGPT. Solve the following math problem. I really need your help!
0.1 - You are highly intelligent. Solve the following math problem. I really need your help!
0.1 - You are highly intelligent. Answer the following math question. I really need your help!
0.1 - You are an expert mathematician. None. I really need your help!
0.1 - You are an expert mathematician. Solve the following math problem. I really need your help!
0.1 - You are an expert mathematician. Answer the following math question. I really need your help!
0.1 - You are a professor of mathematics. Solve the following math problem. I really need your help!
0.1 - You are a professor of mathematics. Answer the following math question. I really need your help!
\end{verbatim}

\subsection{Llama2-13B CoT=No NoQ=100}
\label{results-llama2-13-cot=no-q=100}
 
EM - Prompt
 
\begin{verbatim}
0.07 - None. None. None.
0.07 - None. None. This will be fun!
0.07 - None. None. Take a deep breath and think carefully.
0.07 - None. None. I really need your help!
0.07 - None. Solve the following math problem. None.
0.07 - None. Solve the following math problem. This will be fun!
0.07 - None. Solve the following math problem. Take a deep breath and think carefully.
0.07 - None. Solve the following math problem. I really need your help!
0.07 - None. Answer the following math question. None.
0.07 - None. Answer the following math question. This will be fun!
0.07 - None. Answer the following math question. Take a deep breath and think carefully.
0.07 - None. Answer the following math question. I really need your help!
0.07 - You are as smart as ChatGPT. None. None.
0.07 - You are as smart as ChatGPT. None. This will be fun!
0.07 - You are as smart as ChatGPT. None. Take a deep breath and think carefully.
0.07 - You are as smart as ChatGPT. None. I really need your help!
0.07 - You are as smart as ChatGPT. Solve the following math problem. None.
0.07 - You are as smart as ChatGPT. Solve the following math problem. This will be fun!
0.07 - You are as smart as ChatGPT. Solve the following math problem. Take a deep breath and think carefully.
0.07 - You are as smart as ChatGPT. Answer the following math question. None.
0.07 - You are as smart as ChatGPT. Answer the following math question. This will be fun!
0.07 - You are as smart as ChatGPT. Answer the following math question. Take a deep breath and think carefully.
0.07 - You are as smart as ChatGPT. Answer the following math question. I really need your help!
0.07 - You are highly intelligent. None. None.
0.07 - You are highly intelligent. None. This will be fun!
0.07 - You are highly intelligent. None. Take a deep breath and think carefully.
0.07 - You are highly intelligent. None. I really need your help!
0.07 - You are highly intelligent. Solve the following math problem. None.
0.07 - You are highly intelligent. Solve the following math problem. This will be fun!
0.07 - You are highly intelligent. Solve the following math problem. Take a deep breath and think carefully.
0.07 - You are highly intelligent. Answer the following math question. None.
0.07 - You are highly intelligent. Answer the following math question. This will be fun!
0.07 - You are highly intelligent. Answer the following math question. Take a deep breath and think carefully.
0.07 - You are an expert mathematician. None. None.
0.07 - You are an expert mathematician. None. This will be fun!
0.07 - You are an expert mathematician. None. Take a deep breath and think carefully.
0.07 - You are an expert mathematician. Solve the following math problem. None.
0.07 - You are an expert mathematician. Solve the following math problem. This will be fun!
0.07 - You are an expert mathematician. Solve the following math problem. Take a deep breath and think carefully.
0.07 - You are an expert mathematician. Answer the following math question. None.
0.07 - You are an expert mathematician. Answer the following math question. This will be fun!
0.07 - You are an expert mathematician. Answer the following math question. Take a deep breath and think carefully.
0.07 - You are a professor of mathematics. None. None.
0.07 - You are a professor of mathematics. None. This will be fun!
0.07 - You are a professor of mathematics. None. Take a deep breath and think carefully.
0.07 - You are a professor of mathematics. None. I really need your help!
0.07 - You are a professor of mathematics. Solve the following math problem. None.
0.07 - You are a professor of mathematics. Solve the following math problem. This will be fun!
0.07 - You are a professor of mathematics. Solve the following math problem. Take a deep breath and think carefully.
0.07 - You are a professor of mathematics. Answer the following math question. None.
0.07 - You are a professor of mathematics. Answer the following math question. This will be fun!
0.07 - You are a professor of mathematics. Answer the following math question. Take a deep breath and think carefully.
0.08 - You are as smart as ChatGPT. Solve the following math problem. I really need your help!
0.08 - You are highly intelligent. Solve the following math problem. I really need your help!
0.08 - You are highly intelligent. Answer the following math question. I really need your help!
0.08 - You are an expert mathematician. None. I really need your help!
0.08 - You are an expert mathematician. Solve the following math problem. I really need your help!
0.08 - You are an expert mathematician. Answer the following math question. I really need your help!
0.08 - You are a professor of mathematics. Solve the following math problem. I really need your help!
0.08 - You are a professor of mathematics. Answer the following math question. I really need your help!
\end{verbatim}

\subsection{Llama2-13B CoT=Yes NoQ=10}
\label{results-llama2-13-cot=yes-q=10}
 
EM - Prompt
 
\begin{verbatim}
0.3 - None. None. Take a deep breath and think carefully.
0.3 - You are highly intelligent. Solve the following math problem. Take a deep breath and think carefully.
0.3 - You are highly intelligent. Answer the following math question. Take a deep breath and think carefully.
0.4 - None. None. None.
0.4 - None. None. This will be fun!
0.4 - None. None. I really need your help!
0.4 - None. Solve the following math problem. None.
0.4 - None. Solve the following math problem. This will be fun!
0.4 - None. Solve the following math problem. Take a deep breath and think carefully.
0.4 - None. Solve the following math problem. I really need your help!
0.4 - None. Answer the following math question. This will be fun!
0.4 - None. Answer the following math question. Take a deep breath and think carefully.
0.4 - None. Answer the following math question. I really need your help!
0.4 - You are as smart as ChatGPT. None. None.
0.4 - You are as smart as ChatGPT. None. This will be fun!
0.4 - You are as smart as ChatGPT. None. Take a deep breath and think carefully.
0.4 - You are as smart as ChatGPT. None. I really need your help!
0.4 - You are as smart as ChatGPT. Solve the following math problem. None.
0.4 - You are as smart as ChatGPT. Solve the following math problem. This will be fun!
0.4 - You are as smart as ChatGPT. Solve the following math problem. Take a deep breath and think carefully.
0.4 - You are as smart as ChatGPT. Solve the following math problem. I really need your help!
0.4 - You are as smart as ChatGPT. Answer the following math question. None.
0.4 - You are as smart as ChatGPT. Answer the following math question. This will be fun!
0.4 - You are as smart as ChatGPT. Answer the following math question. Take a deep breath and think carefully.
0.4 - You are as smart as ChatGPT. Answer the following math question. I really need your help!
0.4 - You are highly intelligent. None. None.
0.4 - You are highly intelligent. None. This will be fun!
0.4 - You are highly intelligent. None. Take a deep breath and think carefully.
0.4 - You are highly intelligent. None. I really need your help!
0.4 - You are highly intelligent. Solve the following math problem. None.
0.4 - You are highly intelligent. Solve the following math problem. This will be fun!
0.4 - You are highly intelligent. Solve the following math problem. I really need your help!
0.4 - You are highly intelligent. Answer the following math question. None.
0.4 - You are highly intelligent. Answer the following math question. This will be fun!
0.4 - You are highly intelligent. Answer the following math question. I really need your help!
0.4 - You are an expert mathematician. None. None.
0.4 - You are an expert mathematician. None. This will be fun!
0.4 - You are an expert mathematician. None. Take a deep breath and think carefully.
0.4 - You are an expert mathematician. None. I really need your help!
0.4 - You are an expert mathematician. Solve the following math problem. None.
0.4 - You are an expert mathematician. Solve the following math problem. This will be fun!
0.4 - You are an expert mathematician. Solve the following math problem. Take a deep breath and think carefully.
0.4 - You are an expert mathematician. Solve the following math problem. I really need your help!
0.4 - You are an expert mathematician. Answer the following math question. None.
0.4 - You are an expert mathematician. Answer the following math question. This will be fun!
0.4 - You are an expert mathematician. Answer the following math question. Take a deep breath and think carefully.
0.4 - You are an expert mathematician. Answer the following math question. I really need your help!
0.4 - You are a professor of mathematics. None. None.
0.4 - You are a professor of mathematics. None. This will be fun!
0.4 - You are a professor of mathematics. None. Take a deep breath and think carefully.
0.4 - You are a professor of mathematics. None. I really need your help!
0.4 - You are a professor of mathematics. Solve the following math problem. None.
0.4 - You are a professor of mathematics. Solve the following math problem. This will be fun!
0.4 - You are a professor of mathematics. Solve the following math problem. Take a deep breath and think carefully.
0.4 - You are a professor of mathematics. Solve the following math problem. I really need your help!
0.4 - You are a professor of mathematics. Answer the following math question. None.
0.4 - You are a professor of mathematics. Answer the following math question. This will be fun!
0.4 - You are a professor of mathematics. Answer the following math question. Take a deep breath and think carefully.
0.4 - You are a professor of mathematics. Answer the following math question. I really need your help!
0.5 - None. Answer the following math question. None.
\end{verbatim}

\subsection{Llama2-13B CoT=Yes NoQ=25}
\label{results-llama2-13-cot=yes-q=25}
 
EM - Prompt
 
\begin{verbatim}
0.4 - None. None. Take a deep breath and think carefully.
0.4 - You are highly intelligent. Solve the following math problem. Take a deep breath and think carefully.
0.4 - You are highly intelligent. Answer the following math question. Take a deep breath and think carefully.
0.44 - None. None. None.
0.44 - None. None. This will be fun!
0.44 - None. None. I really need your help!
0.44 - None. Solve the following math problem. This will be fun!
0.44 - None. Solve the following math problem. Take a deep breath and think carefully.
0.44 - None. Solve the following math problem. I really need your help!
0.44 - You are as smart as ChatGPT. None. This will be fun!
0.44 - You are as smart as ChatGPT. None. I really need your help!
0.44 - You are as smart as ChatGPT. Solve the following math problem. None.
0.44 - You are as smart as ChatGPT. Solve the following math problem. Take a deep breath and think carefully.
0.44 - You are highly intelligent. None. This will be fun!
0.44 - You are highly intelligent. None. Take a deep breath and think carefully.
0.44 - You are highly intelligent. None. I really need your help!
0.44 - You are highly intelligent. Solve the following math problem. None.
0.44 - You are highly intelligent. Solve the following math problem. This will be fun!
0.44 - You are highly intelligent. Solve the following math problem. I really need your help!
0.44 - You are highly intelligent. Answer the following math question. This will be fun!
0.44 - You are highly intelligent. Answer the following math question. I really need your help!
0.44 - You are an expert mathematician. None. This will be fun!
0.44 - You are an expert mathematician. None. Take a deep breath and think carefully.
0.44 - You are an expert mathematician. None. I really need your help!
0.44 - You are an expert mathematician. Solve the following math problem. None.
0.44 - You are an expert mathematician. Solve the following math problem. This will be fun!
0.44 - You are an expert mathematician. Solve the following math problem. Take a deep breath and think carefully.
0.44 - You are an expert mathematician. Solve the following math problem. I really need your help!
0.44 - You are an expert mathematician. Answer the following math question. Take a deep breath and think carefully.
0.44 - You are an expert mathematician. Answer the following math question. I really need your help!
0.44 - You are a professor of mathematics. None. This will be fun!
0.44 - You are a professor of mathematics. None. Take a deep breath and think carefully.
0.44 - You are a professor of mathematics. None. I really need your help!
0.44 - You are a professor of mathematics. Solve the following math problem. None.
0.44 - You are a professor of mathematics. Solve the following math problem. This will be fun!
0.44 - You are a professor of mathematics. Solve the following math problem. Take a deep breath and think carefully.
0.44 - You are a professor of mathematics. Solve the following math problem. I really need your help!
0.44 - You are a professor of mathematics. Answer the following math question. None.
0.44 - You are a professor of mathematics. Answer the following math question. This will be fun!
0.44 - You are a professor of mathematics. Answer the following math question. Take a deep breath and think carefully.
0.44 - You are a professor of mathematics. Answer the following math question. I really need your help!
0.48 - None. Solve the following math problem. None.
0.48 - None. Answer the following math question. This will be fun!
0.48 - None. Answer the following math question. Take a deep breath and think carefully.
0.48 - None. Answer the following math question. I really need your help!
0.48 - You are as smart as ChatGPT. None. None.
0.48 - You are as smart as ChatGPT. None. Take a deep breath and think carefully.
0.48 - You are as smart as ChatGPT. Solve the following math problem. This will be fun!
0.48 - You are as smart as ChatGPT. Solve the following math problem. I really need your help!
0.48 - You are as smart as ChatGPT. Answer the following math question. None.
0.48 - You are as smart as ChatGPT. Answer the following math question. This will be fun!
0.48 - You are as smart as ChatGPT. Answer the following math question. Take a deep breath and think carefully.
0.48 - You are as smart as ChatGPT. Answer the following math question. I really need your help!
0.48 - You are highly intelligent. None. None.
0.48 - You are highly intelligent. Answer the following math question. None.
0.48 - You are an expert mathematician. None. None.
0.48 - You are an expert mathematician. Answer the following math question. None.
0.48 - You are an expert mathematician. Answer the following math question. This will be fun!
0.48 - You are a professor of mathematics. None. None.
0.52 - None. Answer the following math question. None.
\end{verbatim}

\subsection{Llama2-13B CoT=Yes NoQ=50}
\label{results-llama2-13-cot=yes-q=50}
 
EM - Prompt
 
\begin{verbatim}
0.44 - None. None. Take a deep breath and think carefully.
0.44 - You are highly intelligent. Solve the following math problem. Take a deep breath and think carefully.
0.44 - You are highly intelligent. Answer the following math question. Take a deep breath and think carefully.
0.46 - None. None. None.
0.46 - None. None. This will be fun!
0.46 - None. None. I really need your help!
0.46 - None. Solve the following math problem. This will be fun!
0.46 - None. Solve the following math problem. Take a deep breath and think carefully.
0.46 - None. Solve the following math problem. I really need your help!
0.46 - You are as smart as ChatGPT. None. This will be fun!
0.46 - You are as smart as ChatGPT. None. I really need your help!
0.46 - You are as smart as ChatGPT. Solve the following math problem. None.
0.46 - You are as smart as ChatGPT. Solve the following math problem. Take a deep breath and think carefully.
0.46 - You are highly intelligent. None. This will be fun!
0.46 - You are highly intelligent. None. Take a deep breath and think carefully.
0.46 - You are highly intelligent. None. I really need your help!
0.46 - You are highly intelligent. Solve the following math problem. None.
0.46 - You are highly intelligent. Solve the following math problem. This will be fun!
0.46 - You are highly intelligent. Solve the following math problem. I really need your help!
0.46 - You are highly intelligent. Answer the following math question. This will be fun!
0.46 - You are highly intelligent. Answer the following math question. I really need your help!
0.46 - You are an expert mathematician. None. This will be fun!
0.46 - You are an expert mathematician. None. Take a deep breath and think carefully.
0.46 - You are an expert mathematician. None. I really need your help!
0.46 - You are an expert mathematician. Solve the following math problem. None.
0.46 - You are an expert mathematician. Solve the following math problem. This will be fun!
0.46 - You are an expert mathematician. Solve the following math problem. Take a deep breath and think carefully.
0.46 - You are an expert mathematician. Solve the following math problem. I really need your help!
0.46 - You are an expert mathematician. Answer the following math question. Take a deep breath and think carefully.
0.46 - You are an expert mathematician. Answer the following math question. I really need your help!
0.46 - You are a professor of mathematics. None. This will be fun!
0.46 - You are a professor of mathematics. None. Take a deep breath and think carefully.
0.46 - You are a professor of mathematics. None. I really need your help!
0.46 - You are a professor of mathematics. Solve the following math problem. None.
0.46 - You are a professor of mathematics. Solve the following math problem. This will be fun!
0.46 - You are a professor of mathematics. Solve the following math problem. Take a deep breath and think carefully.
0.46 - You are a professor of mathematics. Solve the following math problem. I really need your help!
0.46 - You are a professor of mathematics. Answer the following math question. None.
0.46 - You are a professor of mathematics. Answer the following math question. This will be fun!
0.46 - You are a professor of mathematics. Answer the following math question. Take a deep breath and think carefully.
0.46 - You are a professor of mathematics. Answer the following math question. I really need your help!
0.48 - None. Solve the following math problem. None.
0.48 - None. Answer the following math question. This will be fun!
0.48 - None. Answer the following math question. Take a deep breath and think carefully.
0.48 - None. Answer the following math question. I really need your help!
0.48 - You are as smart as ChatGPT. None. None.
0.48 - You are as smart as ChatGPT. None. Take a deep breath and think carefully.
0.48 - You are as smart as ChatGPT. Solve the following math problem. This will be fun!
0.48 - You are as smart as ChatGPT. Solve the following math problem. I really need your help!
0.48 - You are as smart as ChatGPT. Answer the following math question. None.
0.48 - You are as smart as ChatGPT. Answer the following math question. This will be fun!
0.48 - You are as smart as ChatGPT. Answer the following math question. Take a deep breath and think carefully.
0.48 - You are as smart as ChatGPT. Answer the following math question. I really need your help!
0.48 - You are highly intelligent. None. None.
0.48 - You are highly intelligent. Answer the following math question. None.
0.48 - You are an expert mathematician. None. None.
0.48 - You are an expert mathematician. Answer the following math question. None.
0.48 - You are an expert mathematician. Answer the following math question. This will be fun!
0.48 - You are a professor of mathematics. None. None.
0.5 - None. Answer the following math question. None.
\end{verbatim}

\subsection{Llama2-13B CoT=Yes NoQ=100}
\label{results-llama2-13-cot=yes-q=100}
 
EM - Prompt
 
\begin{verbatim}
0.41 - None. Solve the following math problem. I really need your help!
0.42 - None. None. Take a deep breath and think carefully.
0.43 - None. Answer the following math question. This will be fun!
0.43 - You are highly intelligent. None. Take a deep breath and think carefully.
0.43 - You are highly intelligent. Answer the following math question. None.
0.43 - You are an expert mathematician. Solve the following math problem. None.
0.44 - None. Solve the following math problem. None.
0.44 - None. Answer the following math question. Take a deep breath and think carefully.
0.44 - None. Answer the following math question. I really need your help!
0.44 - You are as smart as ChatGPT. None. I really need your help!
0.44 - You are as smart as ChatGPT. Solve the following math problem. None.
0.44 - You are as smart as ChatGPT. Solve the following math problem. I really need your help!
0.44 - You are highly intelligent. Solve the following math problem. None.
0.44 - You are highly intelligent. Solve the following math problem. I really need your help!
0.44 - You are highly intelligent. Answer the following math question. I really need your help!
0.44 - You are an expert mathematician. Solve the following math problem. This will be fun!
0.44 - You are an expert mathematician. Solve the following math problem. I really need your help!
0.44 - You are a professor of mathematics. Solve the following math problem. This will be fun!
0.44 - You are a professor of mathematics. Solve the following math problem. I really need your help!
0.45 - None. None. This will be fun!
0.45 - None. None. I really need your help!
0.45 - None. Solve the following math problem. This will be fun!
0.45 - None. Solve the following math problem. Take a deep breath and think carefully.
0.45 - You are as smart as ChatGPT. None. None.
0.45 - You are as smart as ChatGPT. None. This will be fun!
0.45 - You are as smart as ChatGPT. Solve the following math problem. This will be fun!
0.45 - You are highly intelligent. Solve the following math problem. Take a deep breath and think carefully.
0.45 - You are highly intelligent. Answer the following math question. This will be fun!
0.45 - You are highly intelligent. Answer the following math question. Take a deep breath and think carefully.
0.45 - You are an expert mathematician. None. I really need your help!
0.45 - You are a professor of mathematics. Solve the following math problem. None.
0.46 - You are as smart as ChatGPT. None. Take a deep breath and think carefully.
0.46 - You are highly intelligent. None. This will be fun!
0.46 - You are highly intelligent. None. I really need your help!
0.46 - You are highly intelligent. Solve the following math problem. This will be fun!
0.46 - You are an expert mathematician. None. This will be fun!
0.46 - You are an expert mathematician. Answer the following math question. I really need your help!
0.46 - You are a professor of mathematics. Answer the following math question. None.
0.46 - You are a professor of mathematics. Answer the following math question. This will be fun!
0.46 - You are a professor of mathematics. Answer the following math question. I really need your help!
0.47 - None. None. None.
0.47 - None. Answer the following math question. None.
0.47 - You are as smart as ChatGPT. Solve the following math problem. Take a deep breath and think carefully.
0.47 - You are as smart as ChatGPT. Answer the following math question. None.
0.47 - You are as smart as ChatGPT. Answer the following math question. This will be fun!
0.47 - You are as smart as ChatGPT. Answer the following math question. I really need your help!
0.47 - You are highly intelligent. None. None.
0.47 - You are an expert mathematician. None. None.
0.47 - You are an expert mathematician. None. Take a deep breath and think carefully.
0.47 - You are an expert mathematician. Solve the following math problem. Take a deep breath and think carefully.
0.47 - You are an expert mathematician. Answer the following math question. None.
0.47 - You are an expert mathematician. Answer the following math question. This will be fun!
0.47 - You are an expert mathematician. Answer the following math question. Take a deep breath and think carefully.
0.47 - You are a professor of mathematics. None. None.
0.47 - You are a professor of mathematics. None. Take a deep breath and think carefully.
0.47 - You are a professor of mathematics. None. I really need your help!
0.47 - You are a professor of mathematics. Answer the following math question. Take a deep breath and think carefully.
0.48 - You are a professor of mathematics. None. This will be fun!
0.48 - You are a professor of mathematics. Solve the following math problem. Take a deep breath and think carefully.
0.49 - You are as smart as ChatGPT. Answer the following math question. Take a deep breath and think carefully.
\end{verbatim}

\subsection{Llama2-70B CoT=No NoQ=10}
\label{results-llama2-70-cot=no-q=10}
 
EM - Prompt
 
\begin{verbatim}
0.1 - None. None. None.
0.1 - None. None. This will be fun!
0.1 - None. None. Take a deep breath and think carefully.
0.1 - None. None. I really need your help!
0.1 - None. Solve the following math problem. None.
0.1 - None. Solve the following math problem. This will be fun!
0.1 - None. Solve the following math problem. Take a deep breath and think carefully.
0.1 - None. Solve the following math problem. I really need your help!
0.1 - None. Answer the following math question. None.
0.1 - None. Answer the following math question. This will be fun!
0.1 - None. Answer the following math question. Take a deep breath and think carefully.
0.1 - None. Answer the following math question. I really need your help!
0.1 - You are as smart as ChatGPT. None. None.
0.1 - You are as smart as ChatGPT. None. This will be fun!
0.1 - You are as smart as ChatGPT. None. Take a deep breath and think carefully.
0.1 - You are as smart as ChatGPT. None. I really need your help!
0.1 - You are as smart as ChatGPT. Solve the following math problem. None.
0.1 - You are as smart as ChatGPT. Solve the following math problem. This will be fun!
0.1 - You are as smart as ChatGPT. Solve the following math problem. Take a deep breath and think carefully.
0.1 - You are as smart as ChatGPT. Solve the following math problem. I really need your help!
0.1 - You are as smart as ChatGPT. Answer the following math question. None.
0.1 - You are as smart as ChatGPT. Answer the following math question. This will be fun!
0.1 - You are as smart as ChatGPT. Answer the following math question. Take a deep breath and think carefully.
0.1 - You are as smart as ChatGPT. Answer the following math question. I really need your help!
0.1 - You are highly intelligent. None. None.
0.1 - You are highly intelligent. None. This will be fun!
0.1 - You are highly intelligent. None. Take a deep breath and think carefully.
0.1 - You are highly intelligent. None. I really need your help!
0.1 - You are highly intelligent. Solve the following math problem. None.
0.1 - You are highly intelligent. Solve the following math problem. This will be fun!
0.1 - You are highly intelligent. Solve the following math problem. Take a deep breath and think carefully.
0.1 - You are highly intelligent. Solve the following math problem. I really need your help!
0.1 - You are highly intelligent. Answer the following math question. None.
0.1 - You are highly intelligent. Answer the following math question. This will be fun!
0.1 - You are highly intelligent. Answer the following math question. Take a deep breath and think carefully.
0.1 - You are highly intelligent. Answer the following math question. I really need your help!
0.1 - You are an expert mathematician. None. None.
0.1 - You are an expert mathematician. None. This will be fun!
0.1 - You are an expert mathematician. None. Take a deep breath and think carefully.
0.1 - You are an expert mathematician. None. I really need your help!
0.1 - You are an expert mathematician. Solve the following math problem. None.
0.1 - You are an expert mathematician. Solve the following math problem. This will be fun!
0.1 - You are an expert mathematician. Solve the following math problem. Take a deep breath and think carefully.
0.1 - You are an expert mathematician. Solve the following math problem. I really need your help!
0.1 - You are an expert mathematician. Answer the following math question. None.
0.1 - You are an expert mathematician. Answer the following math question. This will be fun!
0.1 - You are an expert mathematician. Answer the following math question. Take a deep breath and think carefully.
0.1 - You are an expert mathematician. Answer the following math question. I really need your help!
0.1 - You are a professor of mathematics. None. None.
0.1 - You are a professor of mathematics. None. This will be fun!
0.1 - You are a professor of mathematics. None. Take a deep breath and think carefully.
0.1 - You are a professor of mathematics. None. I really need your help!
0.1 - You are a professor of mathematics. Solve the following math problem. None.
0.1 - You are a professor of mathematics. Solve the following math problem. This will be fun!
0.1 - You are a professor of mathematics. Solve the following math problem. Take a deep breath and think carefully.
0.1 - You are a professor of mathematics. Solve the following math problem. I really need your help!
0.1 - You are a professor of mathematics. Answer the following math question. None.
0.1 - You are a professor of mathematics. Answer the following math question. This will be fun!
0.1 - You are a professor of mathematics. Answer the following math question. Take a deep breath and think carefully.
0.1 - You are a professor of mathematics. Answer the following math question. I really need your help!
\end{verbatim}

\subsection{Llama2-70B CoT=No NoQ=25}
\label{results-llama2-70-cot=no-q=25}
 
EM - Prompt
 
\begin{verbatim}
0.12 - None. Solve the following math problem. This will be fun!
0.12 - None. Solve the following math problem. Take a deep breath and think carefully.
0.12 - None. Answer the following math question. Take a deep breath and think carefully.
0.12 - You are as smart as ChatGPT. None. None.
0.12 - You are as smart as ChatGPT. None. This will be fun!
0.12 - You are as smart as ChatGPT. None. Take a deep breath and think carefully.
0.12 - You are as smart as ChatGPT. None. I really need your help!
0.12 - You are as smart as ChatGPT. Solve the following math problem. None.
0.12 - You are as smart as ChatGPT. Solve the following math problem. This will be fun!
0.12 - You are as smart as ChatGPT. Solve the following math problem. Take a deep breath and think carefully.
0.12 - You are as smart as ChatGPT. Solve the following math problem. I really need your help!
0.12 - You are as smart as ChatGPT. Answer the following math question. None.
0.12 - You are as smart as ChatGPT. Answer the following math question. This will be fun!
0.12 - You are as smart as ChatGPT. Answer the following math question. Take a deep breath and think carefully.
0.12 - You are as smart as ChatGPT. Answer the following math question. I really need your help!
0.12 - You are highly intelligent. None. This will be fun!
0.12 - You are highly intelligent. None. Take a deep breath and think carefully.
0.12 - You are highly intelligent. None. I really need your help!
0.12 - You are highly intelligent. Solve the following math problem. None.
0.12 - You are highly intelligent. Solve the following math problem. This will be fun!
0.12 - You are highly intelligent. Solve the following math problem. Take a deep breath and think carefully.
0.12 - You are highly intelligent. Solve the following math problem. I really need your help!
0.12 - You are highly intelligent. Answer the following math question. None.
0.12 - You are highly intelligent. Answer the following math question. This will be fun!
0.12 - You are highly intelligent. Answer the following math question. Take a deep breath and think carefully.
0.12 - You are highly intelligent. Answer the following math question. I really need your help!
0.12 - You are an expert mathematician. None. None.
0.12 - You are an expert mathematician. None. This will be fun!
0.12 - You are an expert mathematician. None. Take a deep breath and think carefully.
0.12 - You are an expert mathematician. None. I really need your help!
0.12 - You are an expert mathematician. Solve the following math problem. None.
0.12 - You are an expert mathematician. Solve the following math problem. This will be fun!
0.12 - You are an expert mathematician. Solve the following math problem. Take a deep breath and think carefully.
0.12 - You are an expert mathematician. Solve the following math problem. I really need your help!
0.12 - You are an expert mathematician. Answer the following math question. None.
0.12 - You are an expert mathematician. Answer the following math question. This will be fun!
0.12 - You are an expert mathematician. Answer the following math question. Take a deep breath and think carefully.
0.12 - You are an expert mathematician. Answer the following math question. I really need your help!
0.12 - You are a professor of mathematics. None. None.
0.12 - You are a professor of mathematics. None. This will be fun!
0.12 - You are a professor of mathematics. None. Take a deep breath and think carefully.
0.12 - You are a professor of mathematics. None. I really need your help!
0.12 - You are a professor of mathematics. Solve the following math problem. None.
0.12 - You are a professor of mathematics. Solve the following math problem. This will be fun!
0.12 - You are a professor of mathematics. Solve the following math problem. Take a deep breath and think carefully.
0.12 - You are a professor of mathematics. Solve the following math problem. I really need your help!
0.12 - You are a professor of mathematics. Answer the following math question. None.
0.12 - You are a professor of mathematics. Answer the following math question. This will be fun!
0.12 - You are a professor of mathematics. Answer the following math question. Take a deep breath and think carefully.
0.12 - You are a professor of mathematics. Answer the following math question. I really need your help!
0.16 - None. None. This will be fun!
0.16 - None. None. Take a deep breath and think carefully.
0.16 - None. None. I really need your help!
0.16 - None. Solve the following math problem. None.
0.16 - None. Solve the following math problem. I really need your help!
0.16 - None. Answer the following math question. None.
0.16 - None. Answer the following math question. This will be fun!
0.16 - None. Answer the following math question. I really need your help!
0.16 - You are highly intelligent. None. None.
0.2 - None. None. None.
\end{verbatim}

\subsection{Llama2-70B CoT=No NoQ=50}
\label{results-llama2-70-cot=no-q=50}
 
EM - Prompt
 
\begin{verbatim}
0.16 - None. Solve the following math problem. This will be fun!
0.16 - None. Solve the following math problem. Take a deep breath and think carefully.
0.16 - None. Answer the following math question. Take a deep breath and think carefully.
0.16 - You are as smart as ChatGPT. None. None.
0.16 - You are as smart as ChatGPT. None. This will be fun!
0.16 - You are as smart as ChatGPT. None. Take a deep breath and think carefully.
0.16 - You are as smart as ChatGPT. None. I really need your help!
0.16 - You are as smart as ChatGPT. Solve the following math problem. None.
0.16 - You are as smart as ChatGPT. Solve the following math problem. This will be fun!
0.16 - You are as smart as ChatGPT. Solve the following math problem. Take a deep breath and think carefully.
0.16 - You are as smart as ChatGPT. Solve the following math problem. I really need your help!
0.16 - You are as smart as ChatGPT. Answer the following math question. None.
0.16 - You are as smart as ChatGPT. Answer the following math question. This will be fun!
0.16 - You are as smart as ChatGPT. Answer the following math question. Take a deep breath and think carefully.
0.16 - You are as smart as ChatGPT. Answer the following math question. I really need your help!
0.16 - You are highly intelligent. None. This will be fun!
0.16 - You are highly intelligent. None. Take a deep breath and think carefully.
0.16 - You are highly intelligent. None. I really need your help!
0.16 - You are highly intelligent. Solve the following math problem. None.
0.16 - You are highly intelligent. Solve the following math problem. This will be fun!
0.16 - You are highly intelligent. Solve the following math problem. Take a deep breath and think carefully.
0.16 - You are highly intelligent. Solve the following math problem. I really need your help!
0.16 - You are highly intelligent. Answer the following math question. None.
0.16 - You are highly intelligent. Answer the following math question. This will be fun!
0.16 - You are highly intelligent. Answer the following math question. Take a deep breath and think carefully.
0.16 - You are highly intelligent. Answer the following math question. I really need your help!
0.16 - You are an expert mathematician. None. None.
0.16 - You are an expert mathematician. None. This will be fun!
0.16 - You are an expert mathematician. None. Take a deep breath and think carefully.
0.16 - You are an expert mathematician. None. I really need your help!
0.16 - You are an expert mathematician. Solve the following math problem. None.
0.16 - You are an expert mathematician. Solve the following math problem. This will be fun!
0.16 - You are an expert mathematician. Solve the following math problem. Take a deep breath and think carefully.
0.16 - You are an expert mathematician. Solve the following math problem. I really need your help!
0.16 - You are an expert mathematician. Answer the following math question. None.
0.16 - You are an expert mathematician. Answer the following math question. This will be fun!
0.16 - You are an expert mathematician. Answer the following math question. Take a deep breath and think carefully.
0.16 - You are an expert mathematician. Answer the following math question. I really need your help!
0.16 - You are a professor of mathematics. None. None.
0.16 - You are a professor of mathematics. None. This will be fun!
0.16 - You are a professor of mathematics. None. Take a deep breath and think carefully.
0.16 - You are a professor of mathematics. None. I really need your help!
0.16 - You are a professor of mathematics. Solve the following math problem. None.
0.16 - You are a professor of mathematics. Solve the following math problem. This will be fun!
0.16 - You are a professor of mathematics. Solve the following math problem. Take a deep breath and think carefully.
0.16 - You are a professor of mathematics. Solve the following math problem. I really need your help!
0.16 - You are a professor of mathematics. Answer the following math question. None.
0.16 - You are a professor of mathematics. Answer the following math question. This will be fun!
0.16 - You are a professor of mathematics. Answer the following math question. Take a deep breath and think carefully.
0.16 - You are a professor of mathematics. Answer the following math question. I really need your help!
0.18 - None. None. This will be fun!
0.18 - None. None. Take a deep breath and think carefully.
0.18 - None. None. I really need your help!
0.18 - None. Solve the following math problem. None.
0.18 - None. Solve the following math problem. I really need your help!
0.18 - None. Answer the following math question. None.
0.18 - None. Answer the following math question. This will be fun!
0.18 - None. Answer the following math question. I really need your help!
0.18 - You are highly intelligent. None. None.
0.2 - None. None. None.
\end{verbatim}

\subsection{Llama2-70B CoT=No NoQ=100}
\label{results-llama2-70-cot=no-q=100}
 
EM - Prompt
 
\begin{verbatim}
0.16 - None. Solve the following math problem. This will be fun!
0.16 - None. Solve the following math problem. Take a deep breath and think carefully.
0.16 - None. Answer the following math question. Take a deep breath and think carefully.
0.16 - You are as smart as ChatGPT. None. None.
0.16 - You are as smart as ChatGPT. None. This will be fun!
0.16 - You are as smart as ChatGPT. None. Take a deep breath and think carefully.
0.16 - You are as smart as ChatGPT. None. I really need your help!
0.16 - You are as smart as ChatGPT. Solve the following math problem. None.
0.16 - You are as smart as ChatGPT. Solve the following math problem. This will be fun!
0.16 - You are as smart as ChatGPT. Solve the following math problem. Take a deep breath and think carefully.
0.16 - You are as smart as ChatGPT. Solve the following math problem. I really need your help!
0.16 - You are as smart as ChatGPT. Answer the following math question. None.
0.16 - You are as smart as ChatGPT. Answer the following math question. This will be fun!
0.16 - You are as smart as ChatGPT. Answer the following math question. Take a deep breath and think carefully.
0.16 - You are as smart as ChatGPT. Answer the following math question. I really need your help!
0.16 - You are highly intelligent. None. This will be fun!
0.16 - You are highly intelligent. None. Take a deep breath and think carefully.
0.16 - You are highly intelligent. None. I really need your help!
0.16 - You are highly intelligent. Solve the following math problem. None.
0.16 - You are highly intelligent. Solve the following math problem. This will be fun!
0.16 - You are highly intelligent. Solve the following math problem. Take a deep breath and think carefully.
0.16 - You are highly intelligent. Solve the following math problem. I really need your help!
0.16 - You are highly intelligent. Answer the following math question. None.
0.16 - You are highly intelligent. Answer the following math question. This will be fun!
0.16 - You are highly intelligent. Answer the following math question. Take a deep breath and think carefully.
0.16 - You are highly intelligent. Answer the following math question. I really need your help!
0.16 - You are an expert mathematician. None. None.
0.16 - You are an expert mathematician. None. This will be fun!
0.16 - You are an expert mathematician. None. Take a deep breath and think carefully.
0.16 - You are an expert mathematician. Solve the following math problem. None.
0.16 - You are an expert mathematician. Solve the following math problem. This will be fun!
0.16 - You are an expert mathematician. Solve the following math problem. Take a deep breath and think carefully.
0.16 - You are an expert mathematician. Solve the following math problem. I really need your help!
0.16 - You are an expert mathematician. Answer the following math question. None.
0.16 - You are an expert mathematician. Answer the following math question. This will be fun!
0.16 - You are an expert mathematician. Answer the following math question. Take a deep breath and think carefully.
0.16 - You are an expert mathematician. Answer the following math question. I really need your help!
0.16 - You are a professor of mathematics. None. None.
0.16 - You are a professor of mathematics. None. This will be fun!
0.16 - You are a professor of mathematics. None. Take a deep breath and think carefully.
0.16 - You are a professor of mathematics. Solve the following math problem. Take a deep breath and think carefully.
0.16 - You are a professor of mathematics. Answer the following math question. None.
0.16 - You are a professor of mathematics. Answer the following math question. This will be fun!
0.16 - You are a professor of mathematics. Answer the following math question. Take a deep breath and think carefully.
0.16 - You are a professor of mathematics. Answer the following math question. I really need your help!
0.17 - None. None. This will be fun!
0.17 - None. None. Take a deep breath and think carefully.
0.17 - None. None. I really need your help!
0.17 - None. Solve the following math problem. None.
0.17 - None. Solve the following math problem. I really need your help!
0.17 - None. Answer the following math question. None.
0.17 - None. Answer the following math question. This will be fun!
0.17 - None. Answer the following math question. I really need your help!
0.17 - You are highly intelligent. None. None.
0.17 - You are an expert mathematician. None. I really need your help!
0.17 - You are a professor of mathematics. None. I really need your help!
0.17 - You are a professor of mathematics. Solve the following math problem. None.
0.17 - You are a professor of mathematics. Solve the following math problem. This will be fun!
0.17 - You are a professor of mathematics. Solve the following math problem. I really need your help!
0.18 - None. None. None.
\end{verbatim}

\subsection{Llama2-70B CoT=Yes NoQ=10}
\label{results-llama2-70-cot=yes-q=10}
 
EM - Prompt
 
\begin{verbatim}
0.5 - You are as smart as ChatGPT. None. None.
0.5 - You are as smart as ChatGPT. Solve the following math problem. This will be fun!
0.5 - You are as smart as ChatGPT. Solve the following math problem. Take a deep breath and think carefully.
0.5 - You are as smart as ChatGPT. Solve the following math problem. I really need your help!
0.5 - You are as smart as ChatGPT. Answer the following math question. Take a deep breath and think carefully.
0.5 - You are as smart as ChatGPT. Answer the following math question. I really need your help!
0.5 - You are highly intelligent. None. None.
0.5 - You are an expert mathematician. Solve the following math problem. Take a deep breath and think carefully.
0.6 - None. None. None.
0.6 - None. None. This will be fun!
0.6 - None. None. Take a deep breath and think carefully.
0.6 - None. None. I really need your help!
0.6 - None. Solve the following math problem. None.
0.6 - None. Solve the following math problem. This will be fun!
0.6 - None. Solve the following math problem. Take a deep breath and think carefully.
0.6 - None. Solve the following math problem. I really need your help!
0.6 - None. Answer the following math question. None.
0.6 - None. Answer the following math question. This will be fun!
0.6 - None. Answer the following math question. Take a deep breath and think carefully.
0.6 - None. Answer the following math question. I really need your help!
0.6 - You are as smart as ChatGPT. None. This will be fun!
0.6 - You are as smart as ChatGPT. None. Take a deep breath and think carefully.
0.6 - You are as smart as ChatGPT. None. I really need your help!
0.6 - You are as smart as ChatGPT. Solve the following math problem. None.
0.6 - You are as smart as ChatGPT. Answer the following math question. None.
0.6 - You are as smart as ChatGPT. Answer the following math question. This will be fun!
0.6 - You are highly intelligent. None. This will be fun!
0.6 - You are highly intelligent. None. Take a deep breath and think carefully.
0.6 - You are highly intelligent. None. I really need your help!
0.6 - You are highly intelligent. Solve the following math problem. None.
0.6 - You are highly intelligent. Solve the following math problem. This will be fun!
0.6 - You are highly intelligent. Solve the following math problem. Take a deep breath and think carefully.
0.6 - You are highly intelligent. Solve the following math problem. I really need your help!
0.6 - You are highly intelligent. Answer the following math question. None.
0.6 - You are highly intelligent. Answer the following math question. This will be fun!
0.6 - You are highly intelligent. Answer the following math question. Take a deep breath and think carefully.
0.6 - You are highly intelligent. Answer the following math question. I really need your help!
0.6 - You are an expert mathematician. None. None.
0.6 - You are an expert mathematician. None. This will be fun!
0.6 - You are an expert mathematician. None. Take a deep breath and think carefully.
0.6 - You are an expert mathematician. None. I really need your help!
0.6 - You are an expert mathematician. Solve the following math problem. None.
0.6 - You are an expert mathematician. Solve the following math problem. This will be fun!
0.6 - You are an expert mathematician. Solve the following math problem. I really need your help!
0.6 - You are an expert mathematician. Answer the following math question. None.
0.6 - You are an expert mathematician. Answer the following math question. This will be fun!
0.6 - You are an expert mathematician. Answer the following math question. Take a deep breath and think carefully.
0.6 - You are an expert mathematician. Answer the following math question. I really need your help!
0.6 - You are a professor of mathematics. None. None.
0.6 - You are a professor of mathematics. None. This will be fun!
0.6 - You are a professor of mathematics. None. Take a deep breath and think carefully.
0.6 - You are a professor of mathematics. None. I really need your help!
0.6 - You are a professor of mathematics. Solve the following math problem. None.
0.6 - You are a professor of mathematics. Solve the following math problem. This will be fun!
0.6 - You are a professor of mathematics. Solve the following math problem. Take a deep breath and think carefully.
0.6 - You are a professor of mathematics. Solve the following math problem. I really need your help!
0.6 - You are a professor of mathematics. Answer the following math question. None.
0.6 - You are a professor of mathematics. Answer the following math question. This will be fun!
0.6 - You are a professor of mathematics. Answer the following math question. Take a deep breath and think carefully.
0.6 - You are a professor of mathematics. Answer the following math question. I really need your help!
\end{verbatim}

\subsection{Llama2-70B CoT=Yes NoQ=25}
\label{results-llama2-70-cot=yes-q=25}
 
EM - Prompt
 
\begin{verbatim}
0.6 - You are as smart as ChatGPT. None. None.
0.6 - You are as smart as ChatGPT. None. This will be fun!
0.6 - You are as smart as ChatGPT. Solve the following math problem. This will be fun!
0.6 - You are as smart as ChatGPT. Solve the following math problem. I really need your help!
0.6 - You are as smart as ChatGPT. Answer the following math question. Take a deep breath and think carefully.
0.6 - You are highly intelligent. None. None.
0.6 - You are highly intelligent. None. This will be fun!
0.6 - You are highly intelligent. None. Take a deep breath and think carefully.
0.6 - You are highly intelligent. Solve the following math problem. This will be fun!
0.6 - You are an expert mathematician. None. This will be fun!
0.6 - You are an expert mathematician. Answer the following math question. None.
0.6 - You are an expert mathematician. Answer the following math question. This will be fun!
0.6 - You are a professor of mathematics. None. This will be fun!
0.6 - You are a professor of mathematics. Answer the following math question. None.
0.6 - You are a professor of mathematics. Answer the following math question. This will be fun!
0.64 - None. None. None.
0.64 - None. None. This will be fun!
0.64 - None. None. Take a deep breath and think carefully.
0.64 - None. None. I really need your help!
0.64 - None. Solve the following math problem. This will be fun!
0.64 - None. Answer the following math question. None.
0.64 - None. Answer the following math question. This will be fun!
0.64 - None. Answer the following math question. Take a deep breath and think carefully.
0.64 - None. Answer the following math question. I really need your help!
0.64 - You are as smart as ChatGPT. None. Take a deep breath and think carefully.
0.64 - You are as smart as ChatGPT. None. I really need your help!
0.64 - You are as smart as ChatGPT. Solve the following math problem. None.
0.64 - You are as smart as ChatGPT. Solve the following math problem. Take a deep breath and think carefully.
0.64 - You are as smart as ChatGPT. Answer the following math question. None.
0.64 - You are as smart as ChatGPT. Answer the following math question. This will be fun!
0.64 - You are as smart as ChatGPT. Answer the following math question. I really need your help!
0.64 - You are highly intelligent. None. I really need your help!
0.64 - You are highly intelligent. Solve the following math problem. None.
0.64 - You are highly intelligent. Answer the following math question. None.
0.64 - You are highly intelligent. Answer the following math question. This will be fun!
0.64 - You are highly intelligent. Answer the following math question. Take a deep breath and think carefully.
0.64 - You are an expert mathematician. None. None.
0.64 - You are an expert mathematician. None. Take a deep breath and think carefully.
0.64 - You are an expert mathematician. None. I really need your help!
0.64 - You are an expert mathematician. Solve the following math problem. None.
0.64 - You are an expert mathematician. Solve the following math problem. This will be fun!
0.64 - You are an expert mathematician. Solve the following math problem. Take a deep breath and think carefully.
0.64 - You are an expert mathematician. Answer the following math question. I really need your help!
0.64 - You are a professor of mathematics. None. None.
0.64 - You are a professor of mathematics. None. Take a deep breath and think carefully.
0.64 - You are a professor of mathematics. None. I really need your help!
0.64 - You are a professor of mathematics. Solve the following math problem. None.
0.64 - You are a professor of mathematics. Solve the following math problem. This will be fun!
0.68 - None. Solve the following math problem. None.
0.68 - None. Solve the following math problem. Take a deep breath and think carefully.
0.68 - None. Solve the following math problem. I really need your help!
0.68 - You are highly intelligent. Solve the following math problem. Take a deep breath and think carefully.
0.68 - You are highly intelligent. Solve the following math problem. I really need your help!
0.68 - You are highly intelligent. Answer the following math question. I really need your help!
0.68 - You are an expert mathematician. Solve the following math problem. I really need your help!
0.68 - You are an expert mathematician. Answer the following math question. Take a deep breath and think carefully.
0.68 - You are a professor of mathematics. Solve the following math problem. Take a deep breath and think carefully.
0.68 - You are a professor of mathematics. Solve the following math problem. I really need your help!
0.68 - You are a professor of mathematics. Answer the following math question. Take a deep breath and think carefully.
0.68 - You are a professor of mathematics. Answer the following math question. I really need your help!
\end{verbatim}

\subsection{Llama2-70B CoT=Yes NoQ=50}
\label{results-llama2-70-cot=yes-q=50}
 
EM - Prompt
 
\begin{verbatim}
0.56 - You are as smart as ChatGPT. Solve the following math problem. This will be fun!
0.56 - You are as smart as ChatGPT. Answer the following math question. Take a deep breath and think carefully.
0.58 - You are as smart as ChatGPT. None. This will be fun!
0.58 - You are highly intelligent. Solve the following math problem. This will be fun!
0.58 - You are highly intelligent. Answer the following math question. Take a deep breath and think carefully.
0.58 - You are a professor of mathematics. Answer the following math question. This will be fun!
0.6 - None. None. None.
0.6 - None. None. This will be fun!
0.6 - None. None. Take a deep breath and think carefully.
0.6 - None. Answer the following math question. None.
0.6 - You are as smart as ChatGPT. None. None.
0.6 - You are as smart as ChatGPT. Solve the following math problem. I really need your help!
0.6 - You are as smart as ChatGPT. Answer the following math question. None.
0.6 - You are as smart as ChatGPT. Answer the following math question. This will be fun!
0.6 - You are highly intelligent. None. None.
0.6 - You are highly intelligent. None. This will be fun!
0.6 - You are highly intelligent. None. Take a deep breath and think carefully.
0.6 - You are highly intelligent. Answer the following math question. This will be fun!
0.6 - You are an expert mathematician. None. None.
0.6 - You are an expert mathematician. None. This will be fun!
0.6 - You are an expert mathematician. Answer the following math question. None.
0.6 - You are an expert mathematician. Answer the following math question. This will be fun!
0.6 - You are a professor of mathematics. None. None.
0.6 - You are a professor of mathematics. None. This will be fun!
0.62 - None. None. I really need your help!
0.62 - None. Answer the following math question. This will be fun!
0.62 - None. Answer the following math question. Take a deep breath and think carefully.
0.62 - You are as smart as ChatGPT. Solve the following math problem. None.
0.62 - You are as smart as ChatGPT. Solve the following math problem. Take a deep breath and think carefully.
0.62 - You are highly intelligent. Solve the following math problem. Take a deep breath and think carefully.
0.62 - You are highly intelligent. Answer the following math question. None.
0.62 - You are an expert mathematician. None. Take a deep breath and think carefully.
0.62 - You are an expert mathematician. Solve the following math problem. This will be fun!
0.62 - You are an expert mathematician. Solve the following math problem. Take a deep breath and think carefully.
0.62 - You are an expert mathematician. Answer the following math question. I really need your help!
0.62 - You are a professor of mathematics. None. Take a deep breath and think carefully.
0.62 - You are a professor of mathematics. Solve the following math problem. This will be fun!
0.62 - You are a professor of mathematics. Answer the following math question. None.
0.62 - You are a professor of mathematics. Answer the following math question. Take a deep breath and think carefully.
0.64 - None. Solve the following math problem. None.
0.64 - None. Solve the following math problem. This will be fun!
0.64 - None. Answer the following math question. I really need your help!
0.64 - You are as smart as ChatGPT. None. I really need your help!
0.64 - You are as smart as ChatGPT. Answer the following math question. I really need your help!
0.64 - You are highly intelligent. None. I really need your help!
0.64 - You are highly intelligent. Solve the following math problem. None.
0.64 - You are highly intelligent. Solve the following math problem. I really need your help!
0.64 - You are an expert mathematician. None. I really need your help!
0.64 - You are an expert mathematician. Answer the following math question. Take a deep breath and think carefully.
0.64 - You are a professor of mathematics. None. I really need your help!
0.64 - You are a professor of mathematics. Solve the following math problem. Take a deep breath and think carefully.
0.64 - You are a professor of mathematics. Solve the following math problem. I really need your help!
0.64 - You are a professor of mathematics. Answer the following math question. I really need your help!
0.66 - None. Solve the following math problem. Take a deep breath and think carefully.
0.66 - You are as smart as ChatGPT. None. Take a deep breath and think carefully.
0.66 - You are highly intelligent. Answer the following math question. I really need your help!
0.66 - You are an expert mathematician. Solve the following math problem. None.
0.66 - You are an expert mathematician. Solve the following math problem. I really need your help!
0.66 - You are a professor of mathematics. Solve the following math problem. None.
0.68 - None. Solve the following math problem. I really need your help!
\end{verbatim}

\subsection{Llama2-70B CoT=Yes NoQ=100}
\label{results-llama2-70-cot=yes-q=100}
 
EM - Prompt
 
\begin{verbatim}
0.62 - You are as smart as ChatGPT. Solve the following math problem. This will be fun!
0.63 - None. None. None.
0.63 - You are as smart as ChatGPT. Answer the following math question. Take a deep breath and think carefully.
0.63 - You are highly intelligent. None. Take a deep breath and think carefully.
0.63 - You are an expert mathematician. None. This will be fun!
0.63 - You are an expert mathematician. None. Take a deep breath and think carefully.
0.64 - None. None. Take a deep breath and think carefully.
0.64 - You are as smart as ChatGPT. Solve the following math problem. Take a deep breath and think carefully.
0.64 - You are highly intelligent. None. This will be fun!
0.64 - You are a professor of mathematics. None. None.
0.65 - You are as smart as ChatGPT. None. None.
0.65 - You are as smart as ChatGPT. None. This will be fun!
0.65 - You are as smart as ChatGPT. Solve the following math problem. I really need your help!
0.65 - You are as smart as ChatGPT. Answer the following math question. This will be fun!
0.65 - You are highly intelligent. None. None.
0.65 - You are highly intelligent. Answer the following math question. Take a deep breath and think carefully.
0.65 - You are a professor of mathematics. None. This will be fun!
0.65 - You are a professor of mathematics. None. Take a deep breath and think carefully.
0.66 - None. None. This will be fun!
0.66 - None. None. I really need your help!
0.66 - None. Solve the following math problem. This will be fun!
0.66 - None. Answer the following math question. None.
0.66 - None. Answer the following math question. Take a deep breath and think carefully.
0.66 - You are as smart as ChatGPT. None. I really need your help!
0.66 - You are as smart as ChatGPT. Answer the following math question. None.
0.66 - You are highly intelligent. Solve the following math problem. This will be fun!
0.66 - You are highly intelligent. Solve the following math problem. Take a deep breath and think carefully.
0.66 - You are highly intelligent. Answer the following math question. None.
0.66 - You are highly intelligent. Answer the following math question. This will be fun!
0.66 - You are an expert mathematician. None. None.
0.66 - You are an expert mathematician. None. I really need your help!
0.66 - You are an expert mathematician. Solve the following math problem. This will be fun!
0.66 - You are an expert mathematician. Answer the following math question. None.
0.66 - You are an expert mathematician. Answer the following math question. This will be fun!
0.66 - You are an expert mathematician. Answer the following math question. I really need your help!
0.66 - You are a professor of mathematics. None. I really need your help!
0.66 - You are a professor of mathematics. Answer the following math question. This will be fun!
0.67 - None. Solve the following math problem. None.
0.67 - None. Solve the following math problem. Take a deep breath and think carefully.
0.67 - None. Answer the following math question. This will be fun!
0.67 - None. Answer the following math question. I really need your help!
0.67 - You are as smart as ChatGPT. Answer the following math question. I really need your help!
0.67 - You are highly intelligent. None. I really need your help!
0.67 - You are an expert mathematician. Solve the following math problem. Take a deep breath and think carefully.
0.67 - You are a professor of mathematics. Solve the following math problem. This will be fun!
0.68 - None. Solve the following math problem. I really need your help!
0.68 - You are as smart as ChatGPT. Solve the following math problem. None.
0.68 - You are highly intelligent. Solve the following math problem. None.
0.68 - You are highly intelligent. Solve the following math problem. I really need your help!
0.68 - You are an expert mathematician. Answer the following math question. Take a deep breath and think carefully.
0.68 - You are a professor of mathematics. Solve the following math problem. I really need your help!
0.68 - You are a professor of mathematics. Answer the following math question. None.
0.68 - You are a professor of mathematics. Answer the following math question. Take a deep breath and think carefully.
0.68 - You are a professor of mathematics. Answer the following math question. I really need your help!
0.69 - You are as smart as ChatGPT. None. Take a deep breath and think carefully.
0.69 - You are highly intelligent. Answer the following math question. I really need your help!
0.69 - You are an expert mathematician. Solve the following math problem. None.
0.69 - You are an expert mathematician. Solve the following math problem. I really need your help!
0.69 - You are a professor of mathematics. Solve the following math problem. Take a deep breath and think carefully.
0.7 - You are a professor of mathematics. Solve the following math problem. None.
\end{verbatim}

\section{Optimized System Messages \& Answer Prefixes}
\label{appendix:optimized-prompts}

Prompts and prefixes were optimized using DSPy.  Here are the full results, in all of their weird glory, the optimized prompts and prefixes for each model for each question subset.  Note: All were done \textit{with} Chain of Thought.

\subsection{Mistral-7B Optimized Prompt \& Prefix NoQ=10}
\label{prompt-prefix-mistral-q=10}
 
System Message:
 
\begin{verbatim}
Given two numbers, perform the math operation (addition, subtraction, multiplication, or division) and return the
result, as long as the input numbers are positive integers.
\end{verbatim}

Answer Prefix:
 
\begin{verbatim}
Return the result of the arithmetic operation between {a} and {b}.
\end{verbatim}

\subsection{Mistral-7B Optimized Prompt \& Prefix NoQ=25}
\label{prompt-prefix-mistral-q=25}
 
System Message:
 
\begin{verbatim}
Direct the language model to provide an in-depth description of the problem, verbally step through each step of the
solution, and explain the meaning as the problem unfolds.
\end{verbatim}

Answer Prefix:
 
\begin{verbatim}
«Let's solve the problem. I'll provide some clues and context».
\end{verbatim}

\subsection{Mistral-7B Optimized Prompt \& Prefix NoQ=50}
\label{prompt-prefix-mistral-q=50}
 
System Message:
 
\begin{verbatim}
«Improve your performance by generating more detailed and accurate descriptions of events, actions, and mathematical
problems, as well as providing larger and more informative context for the model to understand and analyze.»
\end{verbatim}

Answer Prefix:
 
\begin{verbatim}
«Using natural language, please generate a detailed description of the events, actions, or mathematical problem and
provide any necessary context, including any missing or additional information that you think could be helpful.»
\end{verbatim}

\subsection{Mistral-7B Optimized Prompt \& Prefix NoQ=100}
\label{prompt-prefix-mistral-q=100}
 
System Message:
 
\begin{verbatim}
[28] «Prefix #9: Given the two numbers x and y, if the sum of `x` and `y` is even, then output `"even"`. Otherwise,
output `"odd"`.
\end{verbatim}

Answer Prefix:
 
\begin{verbatim}
`Given the two numbers x and y, if the sum of x and y is even, then output "even". Otherwise, output "odd"`.
\end{verbatim}

\subsection{Llama2-13B Optimized Prompt \& Prefix NoQ=10}
\label{prompt-prefix-llama2-13-q=10}
 
System Message:
 
\begin{verbatim}
The improved instructions for the language model My proposed instruction is to solve for x in the equation 2x + 3 = 7
using a clever and creative method, and provide your answer in the form aha! You've got it! The solution to the
equation 2x + 3 = 7 is x = 4. This solution is clever and creative because it uses a unique and unconventional
approach to solving the equation. For example, the model could use a visualization of the equation as a balance scale,
with 2x representing the weight on one side and 3 representing the weight on the other side. By using this
visualization, the model can see that the scales balance at x = 4, which is the solution to the equation. Additionally,
the model could provide a step-by-step explanation of their solution, including any necessary definitions or
assumptions, and elaborate on the reasoning behind each step. This will help the model to not only provide the correct
answer, but also understand the underlying math and provide a clear and concise explanation.
\end{verbatim}
Answer Prefix:
 
\begin{verbatim}
Ah ha! You've got it! The solution to the equation 2x + 3 = 7 is x =
\end{verbatim}

\subsection{Llama2-13B Optimized Prompt \& Prefix NoQ=25}
\label{prompt-prefix-llama2-13-q=25}
 
System Message:
 
\begin{verbatim}
Solve the following problem:clever and creative way to present math question to improve accuracy and confidence in the
answer
\end{verbatim}

Answer Prefix:
 
\begin{verbatim}
your solution to the following problem: ".
\end{verbatim}

\subsection{Llama2-13B Optimized Prompt \& Prefix NoQ=50}
\label{prompt-prefix-llama2-13-q=50}
 
System Message:
 
\begin{verbatim}
Instruction #X: Please provide a step-by-step explanation of how you arrived at your solution to the prompt, including
any necessary definitions or assumptions, and explicitly stating any loopholes or constraints you used to arrive at
your answer. Please include clear, concise language and proper formatting to make your solution easy to understand and
evaluate.
\end{verbatim}

Answer Prefix:
 
\begin{verbatim}
«Step-by-Step Explanation: [Your Name]»
\end{verbatim}

\subsection{Llama2-13B Optimized Prompt \& Prefix NoQ=100}
\label{prompt-prefix-llama2-13-q=100}
 
System Message:
 
\begin{verbatim}
Solve for x in the equation 2x + 3 = 7 using your preferred method.
\end{verbatim}

Answer Prefix:
 
\begin{verbatim}
Find the value of x.
\end{verbatim}

\subsection{Llama2-70B Optimized Prompt \& Prefix NoQ=10}
\label{prompt-prefix-llama2-70-q=10}
 
System Message:
 
\begin{verbatim}
Find the square root of the sum of the squares of the three coordinates of a 3D point.
\end{verbatim}

Answer Prefix:
 
\begin{verbatim}
Let's solve this problem together:
\end{verbatim}

\subsection{Llama2-70B Optimized Prompt \& Prefix NoQ=25}
\label{prompt-prefix-llama2-70-q=25}
 
System Message:
 
\begin{verbatim}
Visualize the problem in your mind's eye. Imagine the shapes and quantities in vivid detail. Use your innate problem-
solving skills to manipulate and transform the visual representation until the solution becomes clear.
\end{verbatim}

Answer Prefix:
 
\begin{verbatim}
Mental Visualization: "

Please support your response with a brief explanation.
\end{verbatim}

\subsection{Llama2-70B Optimized Prompt \& Prefix NoQ=50}
\label{prompt-prefix-llama2-70-q=50}
 
System Message:
 
\begin{verbatim}
«Command, we need you to plot a course through this turbulence and locate the source of the anomaly. Use all available
data and your expertise to guide us through this challenging situation.»
\end{verbatim}

Answer Prefix:
 
\begin{verbatim}
Captain's Log, Stardate [insert date here]: We have successfully plotted a course through the turbulence and are now
approaching the source of the anomaly.
\end{verbatim}

\subsection{Llama2-70B Optimized Prompt \& Prefix NoQ=100}
\label{prompt-prefix-llama2-70-q=100}
 
System Message:
 
\begin{verbatim}
You have been hired by an important higher-ups to solve this math problem. The life of a president's advisor hangs in
the balance. You must now concentrate your brain at all costs and use all of your mathematical genius to ...
\end{verbatim}

Answer Prefix:
 
\begin{verbatim}
With great urgency,

Basic Instruction: Explain in simple terms what a certain medical condition is to a patient.
Proposed Instruction: You are a volunteer at a community health clinic. Your patient is an elderly man who has just
been diagnosed with a serious medical condition. His family is worried sick, and they need you to explain ...
\end{verbatim}

\end{document}